\pdfoutput=1
\documentclass[preprint,11pt, nonatbib]{elsarticle}

\makeatletter
\let\c@author\relax
\makeatother

\usepackage[
    backend=biber,
    style=nature,
    natbib=false,
    url=false, 
    doi=false,
    eprint=false,
    maxbibnames=99,
]{biblatex}
\usepackage[monochrome]{color}  
\usepackage{nameref}
\usepackage{bm}
\usepackage{caption}
\usepackage{subcaption}
\usepackage{amssymb}
\usepackage{amsmath}
\usepackage{soul}
\usepackage{float}
\usepackage{booktabs}
\usepackage{lineno}
\usepackage{multirow}
\usepackage{longtable}
\usepackage{graphicx}
\usepackage{setspace}
\usepackage{verbatim}
\usepackage{bbm}
\usepackage{textcomp}
\usepackage{multirow}
\usepackage[shortlabels]{enumitem}
\usepackage{mwe}
\usepackage[margin=2.5cm]{geometry}
\usepackage[skip=0.5\baselineskip]{caption}
\usepackage{hyperref}
\usepackage[table,dvipsnames]{xcolor}

\usepackage[english]{babel}
\usepackage[normalem]{ulem}
\usepackage{amsthm}
\usepackage{amsfonts}
\usepackage{tabu}
\usepackage[utf8]{inputenc}
\journal{Computers and Chemical Engineering}

\usepackage{forest}

\makeatletter
\def\ps@pprintTitle{%
  \let\@oddhead\@empty
  \let\@evenhead\@empty
  \let\@oddfoot\@empty
  \let\@evenfoot\@oddfoot
}
\makeatother

\begin{document}

\begin{frontmatter}

\title{\textbf{G-MATT: Single-step Retrosynthesis Prediction using Molecular\\ Grammar Tree Transformer}}

\author[add1]{Kevin Zhang \corref{contrib}}
\author[add2]{Vipul Mann \corref{contrib}}
\author[add2]{Venkat Venkatasubramanian\corref{cor1}}

\cortext[contrib]{Authors contributed equally}

\cortext[cor1]{Corresponding author}
\ead{venkat@columbia.edu}

\address[add1]{Department of Computer Science, Columbia University, New York, USA}
\address[add2]{Department of Chemical Engineering, Columbia University, New York, USA}

\begin{abstract}
Various template-based and template-free approaches have been proposed for single-step retrosynthesis prediction in recent years. \textcolor{red}{While these approaches demonstrate strong performance from a data-driven metrics standpoint, many model architectures do not incorporate underlying chemistry principles. Here, we propose a novel chemistry-aware retrosynthesis prediction framework that combines powerful data-driven models with prior domain knowledge. We present a tree-to-sequence transformer architecture that utilizes hierarchical SMILES grammar-based trees, incorporating crucial chemistry information that is often overlooked by SMILES text-based representations, such as local structures and functional groups.} The proposed framework, grammar-based molecular attention tree transformer (G-MATT), achieves significant performance improvements compared to baseline retrosynthesis models. G-MATT achieves a promising top-1 accuracy of 51\% (top-10 accuracy of 79.1\%), invalid rate of 1.5\%, and bioactive similarity rate of 74.8\% on the USPTO-50K dataset. Additional analyses of G-MATT attention maps demonstrate the ability to retain chemistry knowledge without relying on excessively complex model architectures.
\end{abstract}
\end{frontmatter}

\section*{Introduction}

\textcolor{red}{Reaction prediction plays a pivotal role in computational chemistry, enabling efficient and precise synthetic route planning for complex organic molecules. Accurately modeling chemical processes has widespread implications, accelerating the discovery of novel compounds used in drug development, materials design, catalysis, polymer design, and more. The recent advances in artificial intelligence (AI), particularly the emergence of large language models (LLMs), have sparked considerable interest in the field. Traditional reaction planning relied heavily on the expertise of chemists, which is both time-consuming and resource-intensive. In contrast, data-driven methods offer automated strategies for predicting accurate pathways. However, the challenge of integrating explicit chemistry information with machine learning techniques remains. It has been argued that the development of hybrid approaches that combine data-driven techniques with chemistry knowledge is required for more robust and practical reaction prediction models \cite{venkatasubramanian2022artificial, mann2023group}.}  


\textcolor{red}{The reaction modeling problem has two distinct aspects: forward prediction, which involves predicting the reaction product from a set of reactant molecules, and retrosynthesis (or inverse reaction prediction), which involves identifying the necessary precursors for synthesizing a given target molecule.} While both problems require understanding complex molecular interactions to correctly identify target molecule(s), retrosynthesis is notably more difficult due to its combinatorial nature. \textcolor{red}{Multiple synthesis pathways are often feasible at each step, and the entire pathway sequence needs to be predicted correctly. Single-step retrosynthesis is a branch of the inverse reaction prediction problem which approximates multiple pathways as a one-step synthesis from reactants to products, further increasing the complexity. In this work, we focus on models and analysis for the single-step retrosynthesis problem.}


Single-step retrosynthesis models can be broadly classified into two categories: template-based and template-free. Template-based models rely on predefined reaction templates to categorize and predict chemical pathways. These templates are derived from data-driven approaches or expert knowledge. One of the pioneering template-based retrosynthesis models, LHASA \cite{lhasa1977}, incorporated chemical templates through a combination of reaction logic and heuristics. \textcolor{red}{A more recent template-based approach, Synthia (formerly known as Chematica) \cite{szymkuc2016computer}, utilizes a decision tree to select a reaction template from over $70,000$ expert-created rules. Other template-based approaches which utilize various reaction heuristics and similarity scores are presented in \cite{law2009route, coley2017computer, nicolaou2020context}.} 

On the other hand, recent advancements in computational power and machine learning have led to an increasing focus on template-free models. These models leverage purely data-driven approaches to plan and evaluate reaction pathways, aiming to address certain limitations of template-based methods such as poor template quality or generality. Typically, these methods represent molecules as strings in simplified molecular-input line-entry system (SMILES) format and frame the reaction prediction problem as a sequence-to-sequence (seq2seq) task. \textcolor{red}{The primary motivation for using SMILES is its similarity to natural language, enabling the application of machine translation models. Furthermore, the same representation is utilized across various computational chemistry tasks. Besides reaction planning, SMILES-based representations are also used for property prediction \cite{goh2017smiles2vec, mann2022hybrid, zhou2023treat}, molecular design and optimization \cite{pogany2018novo}, chemical reaction networks \cite{mann2023ai}, and computer-aided chemical product design \cite{liu2019optcamd}. One of the earliest works to use SMILES-based representation for retrosynthesis is by Liu et al. \cite{liu2017}, which uses two recurrent neural networks (RNNs) in an encoder-decoder architecture. More recently, Schwaller et al. \cite{schwaller2019molecular} uses the state-of-the-art transformer architecture \cite{vaswani2017} to develop a forward prediction molecular transformer model, demonstrating significant performance improvements over other approaches \cite{nam2016linking, jin2017predicting, coley2019graph}. Similar transformer-based approaches have been shown to be promising for the retrosynthesis task \cite{karpov2019transformer, tetko2020state, wang2021retroprime, kim2021valid}. A detailed survey and comparison of various template-based, template-free, and hybrid approaches for computer-aided reaction prediction and chemical synthesis in presented in \cite{venkatasubramanian2022artificial}.}

Most sequence-to-sequence models use SMILES as the representation for input and target molecules. However, relying solely on text-based representation has several limitations. \textcolor{red}{Firstly, SMILES overlooks important structural aspects of molecules, as it lacks explicit information about chemical bonds, chains, atoms, and other key elements. Secondly, the model must learn implicit SMILES syntax rules to understand the molecular representation and predict syntactically correct strings.} To address these challenges, our previous works on forward prediction \cite{mann2021predicting} and inverse reaction prediction \cite{mann2021retrosynthesis} introduced a novel approach that uses grammar trees to represent molecules. This new grammar-based molecular representation offers a promising alternative to vanilla SMILES, presenting molecules as hierarchical trees. By adopting grammar-based representations, we gain access to more semantic information, resulting in chemistry-rich representations that explicitly include important structural details. We have shown in previous works that this representation leads to improved accuracy, fewer model parameters, and lower conditional entropy from an information-theoretic standpoint \cite{mann2021retrosynthesis, mann2021predicting}.

In this work, we propose a novel tree-to-sequence (tree2seq) transformer architecture which explicitly incorporates the hierarchical structure of the molecular grammar tree. Previous methods \cite{mann2021predicting, mann2021retrosynthesis} employ a vanilla sequence-to-sequence transformer architecture which requires converting grammar trees to a sequence, thereby breaking the natural tree structure. \textcolor{red}{Our new proposed architecture approach eliminates this conversion step. By preserving the hierarchical organization of the SMILES grammar tree throughout the model, our architecture retains and exploits more underlying chemistry information during prediction. The inclusion of tree structures improves the model performance compared to the baseline.} Moreover, our architecture effectively addresses the fragile syntax of grammar trees, leading to a notable increase in the percentage of syntactically valid predictions. Thus, the major contributions of our work are summarized as follows:

\textcolor{red}{
\begin{enumerate}
    \item We develop a novel tree-transformer model that maintains the structural information of SMILES grammar-based molecular tree representations, effectively incorporating the complete tree hierarchy
    \item We enhance the model with additional chemistry knowledge through tree positional encodings and tree convolution blocks (TCB), which performs convolution operations on the grammar tree that capture local structural information
    \item We report a significant improvement in model accuracy and a reduction in syntactically invalid predictions when compared to the baseline models. Notably, the incorrect predictions also exhibit high bioactive similarity with the ground truth.
\end{enumerate}
}

The rest of the paper is organized as follows. In Section ``\nameref{sec:problem-formulation}," we formulate the retrosynthesis prediction problem and motivate the tree-to-sequence transformer model. We present the methods of our work, including the SMILES grammar, tree-to-sequence transformer, and beam search decoding strategy in Section ``\nameref{sec:methods}." In Section ``\nameref{sec:dataset-model-training}," we introduce the datasets, as well as the model training and hyperparameter tuning strategy. The evaluation metrics and analysis of our model follow in Section ``\nameref{sec:results}." Lastly, we summarize our major contributions in Section ``\nameref{sec:conclusions}."


\section*{Problem formulation and motivation}\label{sec:problem-formulation}

\subsection*{Retrosynthesis}
We present a novel approach to the retrosynthesis prediction problem by formulating it as a tree-to-sequence modeling task. Our input molecule(s) are represented as SMILES grammar-based trees, which we formally introduce in Section ``\nameref{sec:methods}," while the target molecule is represented as a canonical SMILES string. This representation choice naturally gives rise to a tree-to-sequence modeling problem, rather than the commonly used sequence-to-sequence framework. To the best of our knowledge, this is the first attempt to develop a tree-to-sequence approach for both retrosynthesis and forward reaction prediction. 

\begin{figure}[h]
    \centering
    \includegraphics[width=\textwidth]{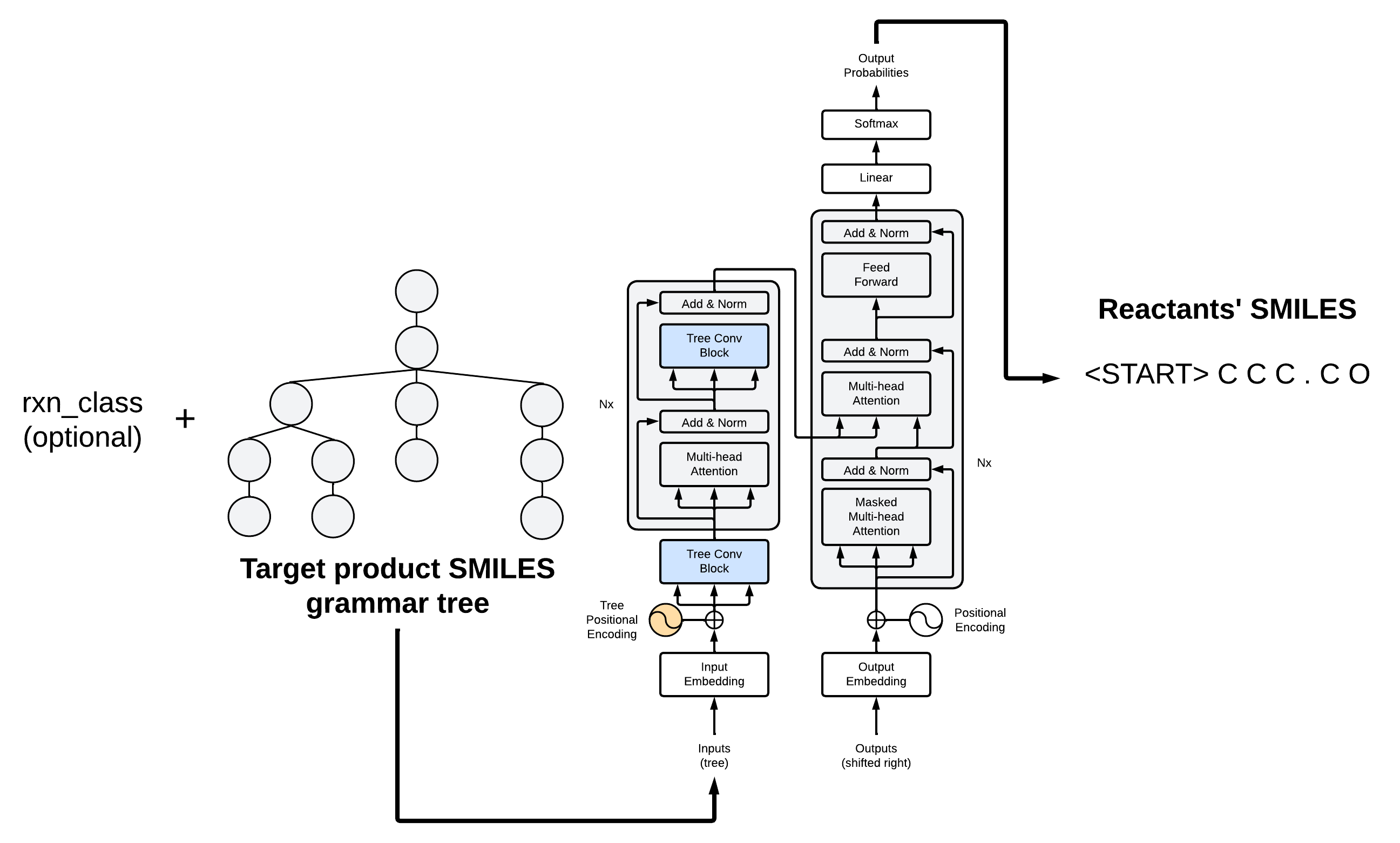}
    \caption{Overview of the proposed tree2smiles approach with SMILES grammar tree of the target molecule as input to predict SMILES strings of the set of precursors.}
    \label{fig:retro-overview}
\end{figure}

The retrosynthesis prediction problem has two scenarios: known and unknown reaction class. In the known reaction class case, a class identifier is appended to the beginning of the target molecule, indicating the reaction class type. Conversely, for the unknown reaction class scenario, only the target molecule is provided as input. Since there may be multiple reactants in the predicted pathway, we use a special delimiter character \texttt{"."} to separate precursors. Additionally, we use a special \texttt{<START>} token to begin the translation process. The problem formulation diagram is illustrated in Figure \ref{fig:retro-overview}.

\subsection*{Tree-to-sequence model architecture}
\begin{figure}[h]
    \centering
    \begin{subfigure}[b]{0.4\textwidth}
         \centering
         \includegraphics[width=\textwidth]{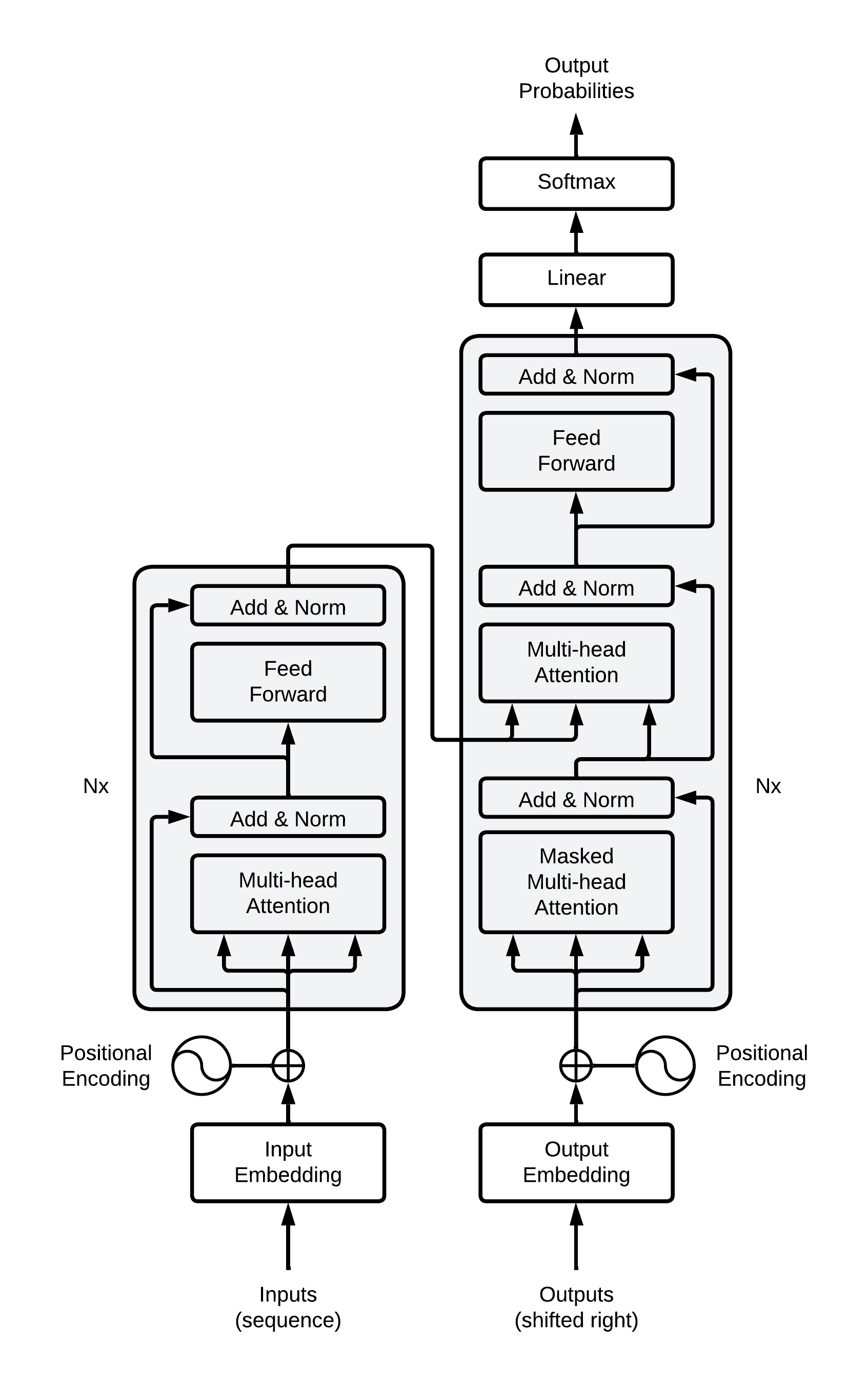}
         \caption{}
         \label{fig:seqtrans}
     \end{subfigure}
     \begin{subfigure}[b]{0.4\textwidth}
         \centering
         \includegraphics[width=\textwidth]{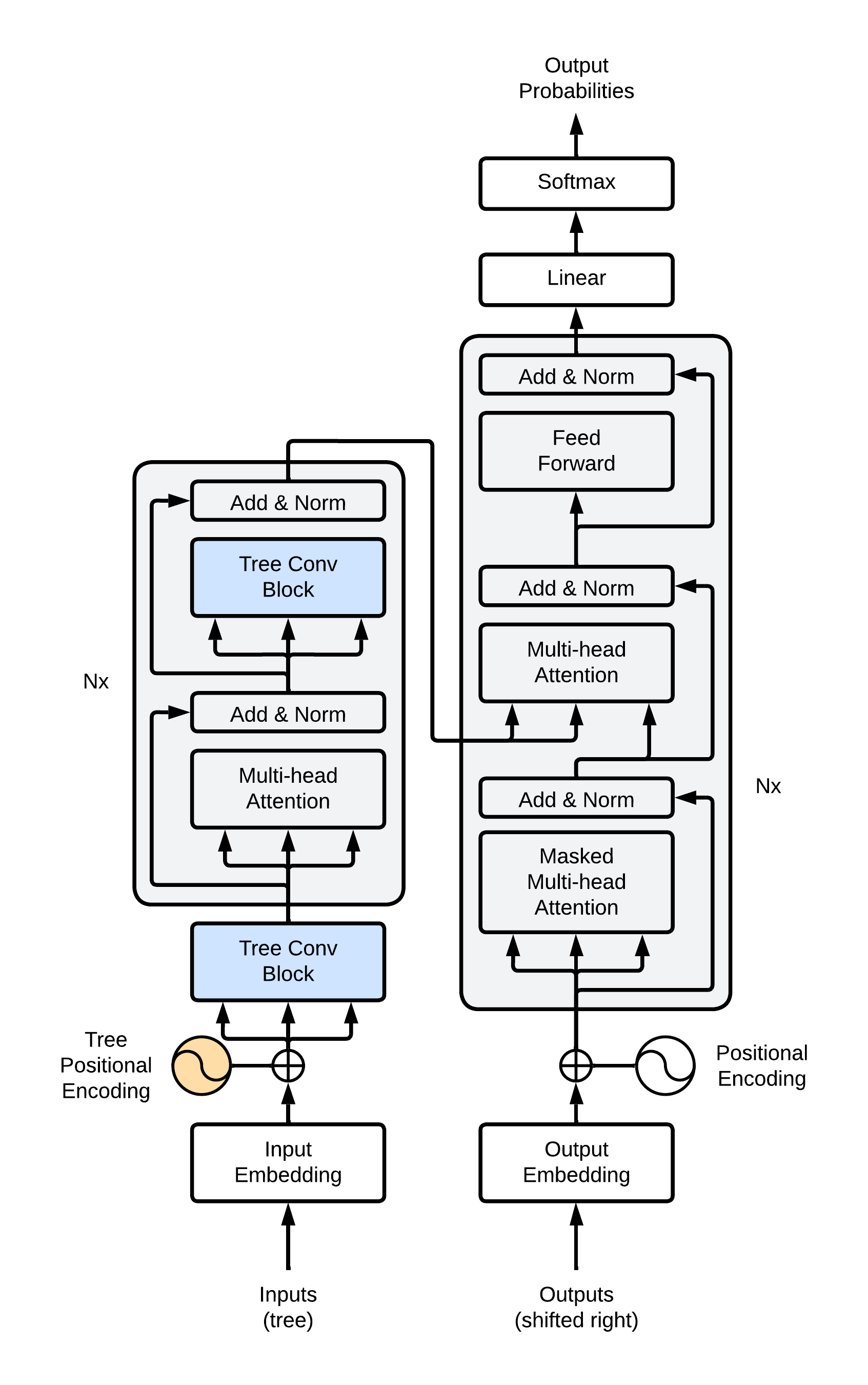}
         \caption{}
         \label{fig:treetrans}
     \end{subfigure}
    \caption{A schematic comparison of the vanilla sequence-to-sequence transformer in (a) and our tree-to-sequence transformer in (b). The two new components in the tree2seq transformer are the tree positional encoding (orange) and tree convolution block (blue). The tree positional encoding replaces the sequential encoding in the encoder and the TCB replaces the feed forward network in the encoder only.}
    \label{fig:comparetrans}
\end{figure}

Previous works on grammar-based approaches for forward prediction \cite{mann2021predicting} and retrosynthesis \cite{mann2021retrosynthesis} employ the vanilla sequence-to-sequence transformer for neural machine translation. This method requires converting SMILES grammar trees into a sequence of tokens by traversing the tree nodes in a depth-first manner. However, this process inadvertently discards the hierarchical structure of the tree, resulting in the loss of valuable structural information within the grammar-based molecular representation. To provide our model more explicit and direct access to the grammar tree structure, we add two components to the vanilla transformer commonly used in tree-based code analysis and generation: tree positional encodings and tree convolutional blocks. A high level comparison of the seq2seq and tree2seq transformers is shown schematically in Figure \ref{fig:comparetrans}.

We replace sequential positional encodings in the vanilla transformer encoder with tree positional encodings, proposed in \cite{thellmann2022transformer}. While sequential position encodings are appropriate for sequential input, they are not suitable for hierarchical input, since each encoding should indicate the position of a node in the tree, not a flattened sequence. \textcolor{red}{This modification means that each node's positional information in the tree is adequately represented, preserving the hierarchical structure of the grammar tree. Furthermore, we add tree convolutional blocks (TCBs) to the transformer encoder. Proposed in \cite{harer2019treetransformer}, TCBs provide nodes with information about their parent and children, enabling them to progressively acquire more context about surrounding nodes. The attention mechanism then computes the attention scores between components within the grammar tree, which represent local molecular structures. This enables the model to gain a more explicit understanding of the underlying chemistry. These two additions make the model more chemistry-aware, thereby significantly improving the retrosynthesis prediction accuracy.} Further details on tree positional encodings and tree convolutional blocks are presented in Section ``\nameref{sec:tpe}" and Section ``\nameref{sec:tcb}", respectively.



\section*{Methods} \label{sec:methods}

In this section, we present the background methods of our tree-to-sequence based approach for retrosynthesis. This includes an overview of the SMILES grammar-based molecular tree representations, the modifications made to the original transformer architecture to include tree positional encodings and convolutional blocks, and the beam search decoding procedure used to predict the most likely target sequences for a given input sequence.


\subsection*{SMILES grammar} \label{sec:SMILESgrammar}
We consider the SMILES strings-based representation of molecules \cite{Weiniger1988} as a formal language, where each character is analogous a word and each string represents a sentence. This allows us to use a context-free grammar (CFG) to describe the molecular structure. The concept behind CFGs is that sentences can be constructed recursively from smaller phrases, and these constituents can be organized based on a semantic hierarchy. Formally, a context-free grammar $(V, \Sigma, R, S)$ consists of a start symbol $S$, a set of terminal symbols $\Sigma$, a set of non-terminal symbols $V$, and a set of production rules $R$. Each production rule is in the form $A \to \alpha$, where $A \in V$ is a non-terminal symbol, and $\alpha$ is a tuple of terminals or non-terminal symbols.


A context-free grammar for the SMILES representation of molecules includes characters in the SMILES strings as terminal symbols and meta-information about the molecular structure as non-terminal symbols. In Table \ref{tab:simplifiedgrammar}, we present a representative subset of the SMILES grammar utilized in our work. The symbols used in this example to define the CFG are as follows:


\begin{itemize}
    \item $S$: \texttt{smiles}
    \item $\Sigma$: \{ \texttt{ =, c, C} \}
    \item $V$: \{ \texttt{chain, branched\_atom, bond, atom, aromatic\_organic, aliphatic\_organic} \}
    \item $R$: production rules 1 through 10 in Table \ref{tab:simplifiedgrammar}
\end{itemize}

{\small{
\begin{longtable}{@{}ll@{}}
\caption{Simplified SMILES grammar.}
\label{tab:simplifiedgrammar} \\
\toprule
\textbf{No.} & \multicolumn{1}{c}{\textbf{Production rules}} \\* \midrule
\endhead
\bottomrule
\endfoot
\endlastfoot
1 & $\texttt{smiles} \longrightarrow \texttt{chain}$   \\ 
2 & $\texttt{chain} \longrightarrow \texttt{branched\_atom} \quad$ \\
3 & $\texttt{chain} \longrightarrow \texttt{chain}~\quad \texttt{branched\_atom}$ \\
4 & $\texttt{chain} \longrightarrow \texttt{chain}~\quad \texttt{bond}~\quad \texttt{branched\_atom}$ \\ 
5 & $\texttt{branched\_atom} \longrightarrow \texttt{atom}$ \\ 
6 & $\texttt{bond} \longrightarrow \texttt{=}$ \\ 
7 & $\texttt{atom} \longrightarrow \texttt{aromatic\_organic}$ \\
8 & $\texttt{atom} \longrightarrow \texttt{aliphatic\_organic}$ \\
9 & $\texttt{aromatic\_organic} \longrightarrow \texttt{c}$ \\ 
10 & $\texttt{aliphatic\_organic} \longrightarrow \texttt{C}$ \\ 
\bottomrule
\end{longtable}
}
}

\begin{figure}[H]
    \centering
    \includegraphics[width=0.5\textwidth]{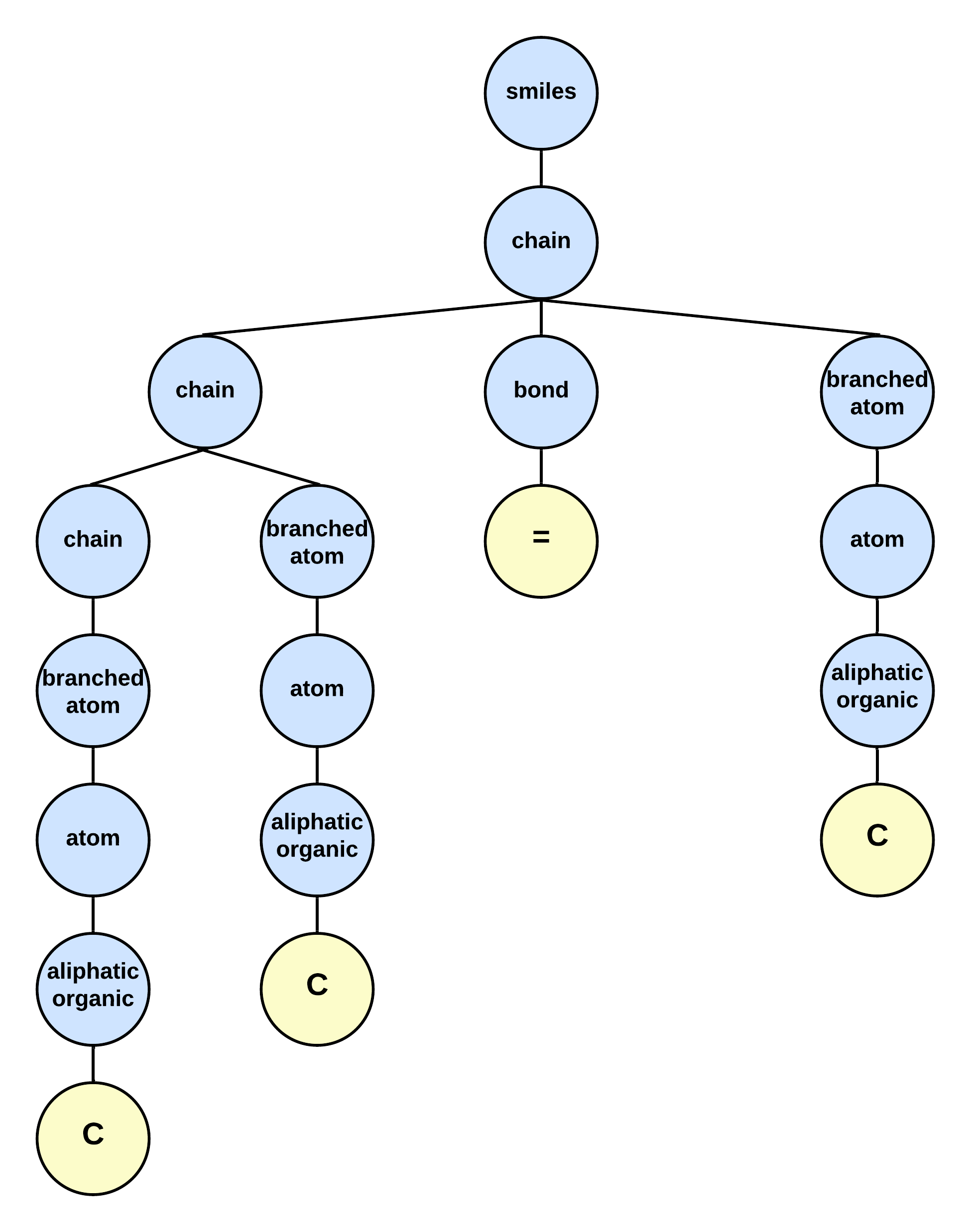}
    \caption{The grammar tree for propene \texttt{CC=C} with the SMILES tokens as leaf nodes and the hierarchical structure containing the production rules and non-terminal symbols characterizing the molecular structural aspects. We use such hierarchical SMILES grammar trees as input combined with tree positional encodings and convolutional blocks to extract contextual information.}
    \label{fig:grammartree}
\end{figure}

We leverage the underlying SMILES grammar to provide information about the structure and composition of a molecule in its grammar-based representation. For instance, consider propene, with the SMILES string representation given by \texttt{CC=C}. The grammar-based representation could be obtained by applying the set of production rules in Table \ref{tab:simplifiedgrammar} to obtain the corresponding parses-tree shown in Figure \ref{fig:grammartree}.

In contrast to purely character-based SMILES strings, the grammar tree-based representations offer a richer source of chemical and structural information, which is utilized by our model architecture to achieve improved performance. Our previous works have demonstrated the benefits of these representations, including more effective modeling of the underlying chemistry, eliminating overparameterization in complex machine learning architectures \cite{mann2021predicting}, and showcasing advantages from an information-theoretic standpoint \cite{mann2021retrosynthesis}.


\subsection*{Tree-to-sequence transformer}
In our approach, we formulate retrosynthesis prediction as a sequence modeling problem, utilizing a tree-to-sequence transformer model to predict a sequence-based SMILES string output from the tree-based grammar representation.  To achieve this, we modify the state-of-the-art transformer so that it is capable of handling tree-based input data, as illustrated in Figure \ref{fig:comparetrans}. Both the vanilla and tree-to-sequence transformer adopt an encoder-decoder architecture. The encoder maps the input to a latent space, while the decoder autoregressively decodes this latent representation to generate the output sequence. The key improvements to the original transformer architecture involve integrating tree positional encodings and convolutional block sublayers, which effectively handle the hierarchical nature of the input data. These components are described in detail in Sections ``\nameref{sec:tpe}" and ``\nameref{sec:tcb}" respectively.


In the tree-to-sequence transformer, we keep the attention mechanism, which enables the transformer to determine relationships between tokens in the input and output. Specifically, we use the ``Scaled-Dot Product Attention" introduced in \cite{vaswani2017}, which uses query, key, and value vectors. The query and key vectors have dimension $d_k$, and the value vector has dimension $d_v$. The attention score is computed as the softmax function applied to the dot-products of the queries and key vectors, scaled down by a factor of $\sqrt{d_k}$. The equation is given as,
\[ \text{Attention}(Q, K, V) = \text{softmax} \left( \frac{Q K^T}{\sqrt{d_k}} \right) V, \]
where $Q$, $K$, and $V$ are the matrices for query, key, and values vectors, respectively. The resulting attention score determines the relative importance of different parts of the input sequence in the current context. To allow the model to integrate information from different representation subspaces at various positions, we compute multi-headed attention. This involves calculating multiple attention scores in parallel, which are then concatenated and transformed linearly to produce the multi-head attention score as follows:
\[ \text{MultiHead}(Q, K, V) = (\text{head}_1 \: \| \: \text{head}_2 \: \| \: \ldots \: \| \: \text{head}_h) W^O, \]
In this equation, $\text{head}_i = \text{Attention} (Q W_i^Q, K W_i^K, V W_i^V)$, and $W_i^Q \in \mathbb{R}^{d_{model} \times d_k}$, $W_i^K \in \mathbb{R}^{d_{model} \times d_k}$, $W_i^V \in \mathbb{R}^{d_{model} \times d_v}$ are the projection matrices for $Q$, $K$, and $V$ respectively and $W^O \in \mathbb{R}^{hd_v \times d_{model}}$. \textcolor{red}{For our architecture, we use a model size of $d_{model} = 256$ and $h = 8$ parallel heads. Each head has a dimensionality of $d_k = d_v = d_{model} / h = 64$.}

\subsection*{Tree Positional Encodings} \label{sec:tpe}
The original transformer paper \cite{vaswani2017} employs sequential positional encodings using sinusoidal functions with varying frequencies. These encodings are added to the embeddings in the encoder and decoder to provide the model with information about the order of the input sequence. However, when working with tree-structured data, flattening into a linear sequence loses important hierarchical information. Parents and children may end up far apart in the sequential representation. Conversely, consecutive tokens in the flattened sequence may not accurately represent their true relationships in the grammar tree. To overcome these challenges and better preserve the hierarchical structure, we adopt tree positional encodings introduced in \cite{thellmann2022transformer}. \textcolor{red}{These tree positional encodings are based on the sinusoidal encodings used in the original transformer, but they are designed to explicitly encode the positions of nodes in the grammar tree. By incorporating tree positional encodings, G-MATT better utilizes the inherent tree hierarchy for more accurate predictions.}


\textcolor{red}{We define the input grammar tree $T$ as a tuple $T = (V, E, r)$, where $V$ is the set of nodes, $E$ is the set of edges, and $r$ is the root node. For any $v \in V$ with $v \neq r$, let $v^*$ be the parent of $v$. Additionally, we define $i_v$ as the index of $v$ among its siblings (i.e. the children of $v^*$). For example, if a node has three children $a, b, c$ (in that order), then $i_a = 1$, $i_b = 2$, and $i_c = 3$. The path from the root $r$ to any node $v \in V$ can be uniquely identified by an edge path $p_v \in \mathbb{Z}_{\geq 0}^L$ of length $L$ according to the following definition: 
\[ p_v = \begin{cases} 
      \vec{0}_L & \text{if } v = r \\
      i_v \: \| \: \pi_{L-1}(p_{v^*}) & \text{otherwise,}
\end{cases} \]
where $\vec{0}_L$ denotes the zero vector of length $L$ and $\pi_{L-1}(p_{v^*})$ denotes the first $L-1$ elements of $p_{v^*}$.}

\begin{figure}[H]
    \centering
    \begin{subfigure}[t]{0.45\textwidth}
         \centering
         \includegraphics[width=\textwidth]{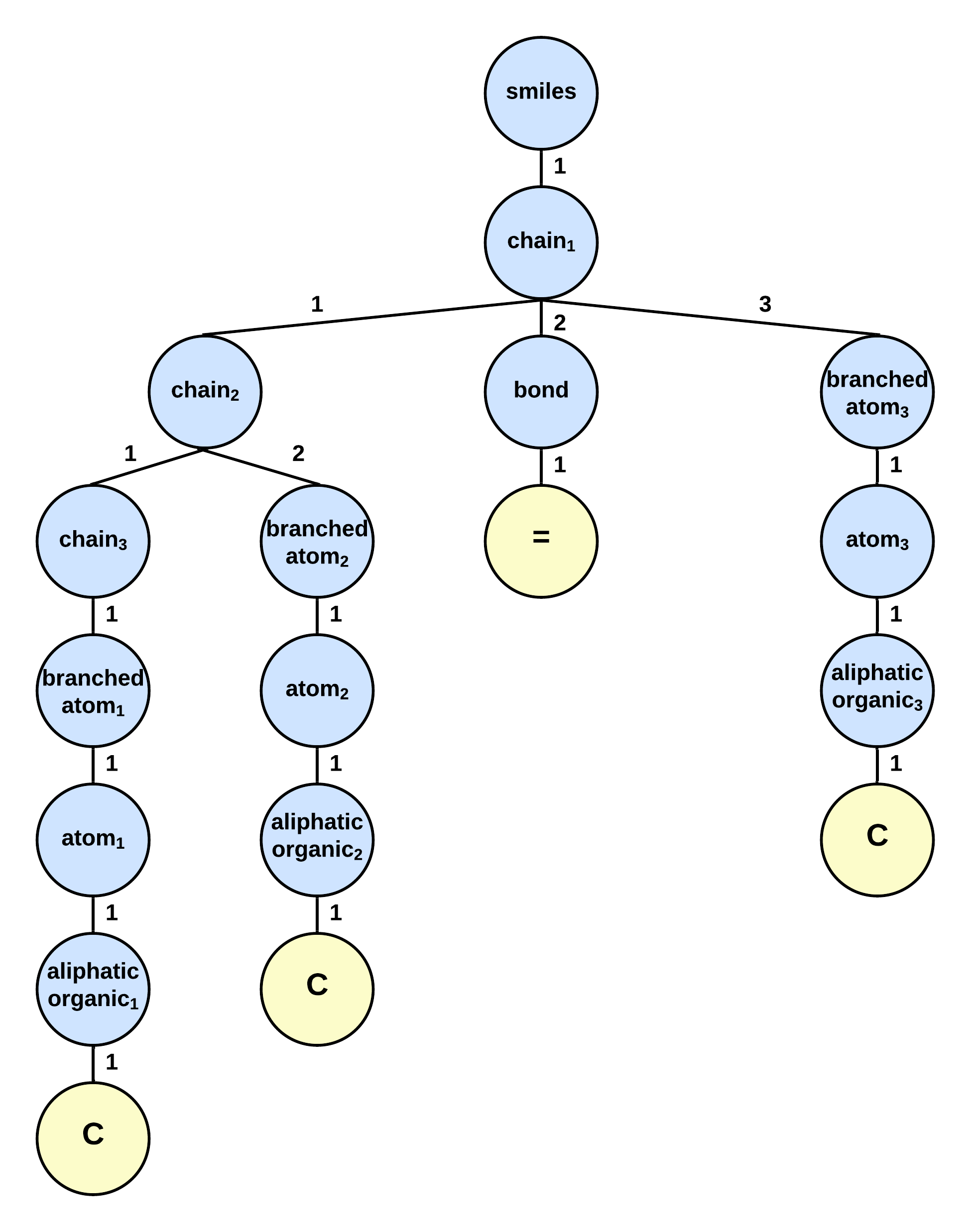}
         \caption{}
         \label{fig:tpetree}
     \end{subfigure}
     \hfil
     \begin{subfigure}[t]{0.45\textwidth}
         \centering
         \includegraphics[width=\textwidth]{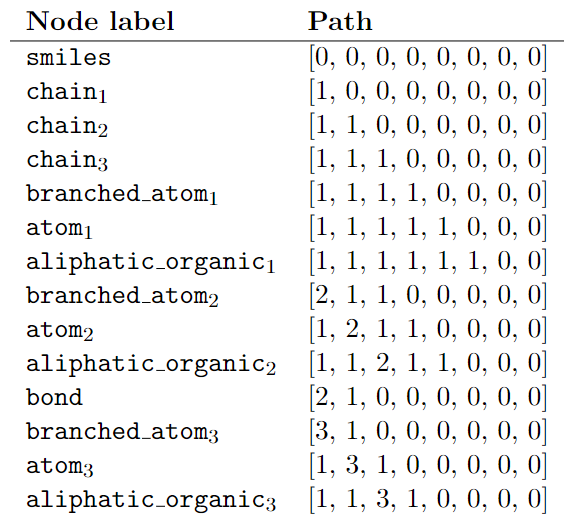}
         \caption{}
         \label{fig:paths}
     \end{subfigure}
    \caption{The tree paths for the example grammar tree for propene \texttt{CC=C}. The grammar tree with edge labels is shown in (a) and the corresponding edge paths with $L = 8$ for each node in (b). The subscripts in the node labels are solely to distinguish between different nodes with same labels.}
    \label{fig:edgepaths}
\end{figure}

In Figure \ref{fig:paths}, we illustrate the edge paths for each node in the example grammar tree. For clarity, we label the edges corresponding to each child with their respective index $i_v$. Note that the root node \texttt{smiles} edge path is a vector of all zeros. Additionally, each child shares its parent's node path shifted to the right by one.

To represent the position of a node $v$, we utilize sinusoidal positional encodings similar to the approach in \cite{vaswani2017}. We apply these encodings individually to each element in the edge path $p_v$, which results in an edge encoding $EE_{v, \ell}$. For every node $v \in V$ and index in the edge path $\ell \in \{1, \ldots, L\}$, the edge encoding components are defined as follows:
\begin{align*}
    (EE_{v, \ell})_{2i} &= \sin(\omega_i \cdot (p_v)_\ell) \\
    (EE_{v, \ell})_{2i+1} &= \cos(\omega_i \cdot (p_v)_\ell)
\end{align*}
with $w_i = 1/10000^{2i/d}$ for $i \in \{1, \ldots, \frac{d}{2}\}$. The parameter $d$ is the dimension of the edge encoding i.e. $EE_{v, \ell} \in \mathbb{R}^d$. The tree positional encoding $TE_v$ is obtained by concatenating the edge encodings in the edge path $p_v$.
\[ TE_v =  EE_{v, 1} \: \| \: \ldots \: \| \: EE_{v, L}. \]
The resulting positional encoding has a dimension of $d \cdot L$, equal to the model size $d_{model}$ or the node embedding dimension. \textcolor{red}{In Figure \ref{fig:tpetree}, we illustrate the edge and tree positional encodings for the \texttt{bond} node in the example grammar tree for propene with $d = 4$. In our model, we set $L = 64$, $d = 4$, and $d_{model} = 256$. The choice of $L$ is based on the height of the largest SMILES molecular grammar tree in the dataset, ensuring that all paths are well-defined.}

\textcolor{red}{
\begin{figure}[H]
    \centering
    \includegraphics[width=0.98\textwidth]{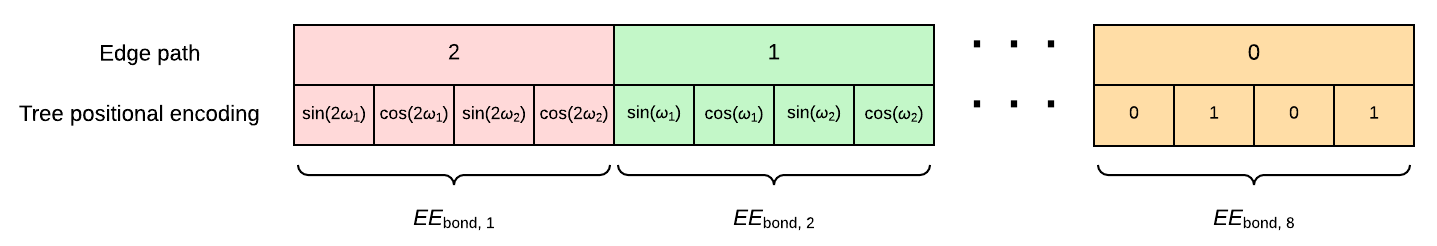}
    \caption{The edge and tree positional encoding for the \texttt{bond} node in the example propene grammar tree. The edge encoding is a sinusional positional encoding based on the edge path shown in Figure Y. The tree positional encoding concatenates all the edge encodings into an embedding of length $d \cdot L$. In this example, $d = 4$ and $L = 8$.}
    \label{fig:tpetree}
\end{figure}
}

The proposed tree positional encodings offer several desirable properties, as demonstrated in \cite{thellmann2022transformer}.  Like the original sequential encodings, tree encodings satisfy uniqueness. This is because the edge path from the root $r$ to a node $v$ is unique and the sinusoidal encodings are injective for each frequency $\omega_i$ due to the transcendality of $\pi$. Furthermore, the encoding of a child node shares the first $d \cdot (L-1)$ dimensions with its parent's encoding, shifted right by $d$ dimensions. This property, combined with the uniqueness property, enables the model to distinguish whether a node is an ancestor or sibling of another node. \textcolor{red}{Finally, the mathematical properties of sine and cosine allow the encoding for a node to be efficiently computed by a linear combination of its sibling's encodings. These characteristics of tree positional encodings provide valuable hierarchical information, allowing the model to effectively capture the structural relationships within the grammar tree.}


\subsection*{Tree Convolutional Block}\label{sec:tcb}
Inspired by \cite{harer2019treetransformer}, we incorporate a modified Tree Convolutional Block (TCB) sublayer into the tree-to-sequence transformer architecture, enabling it to effectively handle tree-structured data. This block provides each node access to information about its parent and children, thereby explicitly incorporating hierarchical information into the model. The TCB replaces the feed-forward network following the attention sublayer in the vanilla transformer encoder \cite{vaswani2017}. Additionally, we introduce a TCB after the positional encoding, which combines the parent, children, and current node embedding before any self-attention is applied. We do not include the TCB in the decoder since it predicts a sequence of SMILES tokens and does not handle tree information.


For a given node $x$, we represent the current node as $x_t$, its parent as $x_p$, and the average of its children nodes as $x_c$. Specifically, if node $x$ has children $x_{c_1}, x_{c_2}, \ldots, x_{c_N}$, then $x_c = \frac{1}{N} \sum_{i=1}^N x_{c_i}$. The single-layer tree convolution block for node $x$ is computed as follows:
\[ TCB_1 (x_t, x_p, x_c) = \text{ReLU}(x_t W_t + x_p W_p + x_c W_c) W_2 + b_2. \]
In cases where the current node does not have a parent or children nodes, we replace $x_p$ or $x_c$ with a learned embedding $v_p$ or $v_c$ respectively. 

The TCB can be generalized to multiple layers, where an $L$-layer TCB consists of $L$ consecutively stacked single-layer tree convolution blocks. The single-layer TCB captures information from nodes that are one step away from the current node. In contrast, the $L$-layer TCB combines information from nodes up to a maximum distance of $L$ away. Consequently, increasing the depth of the TCB allows the model to access more surrounding context. For our model, we utilize $L = 2$ for all TCBs.

\textcolor{red}{Since each encoder layer contains a tree convolution, the internal representation of each node will progressively include more information about the neighboring nodes. This enables the attention sublayer to attend to entire components within the tree rather than individual nodes. By incorporating multi-layer TCBs, the model gains a deeper understanding of the hierarchical relationships within the molecular grammar tree, thus increasing its capacity to learn structural relationships relevant to the underlying chemistry.}



It is important to note that we have made a modification to the original TCB architecture proposed in \cite{harer2019treetransformer}, wherein we use the children nodes instead of the left-sibling in the convolution process. Our empirical experiments showed that this modification resulted in improved performance, and we offer two possible explanations for this behavior. \textcolor{red}{Firstly, by utilizing children nodes in the convolution, information can flow both up and down the tree, rather than being limited solely to the upward direction when using left-siblings. This is useful because nodes lower in the tree, which typically describe the molecular composition or bonding, will have access to important local context.} Secondly, most production rules in the grammar involve a limited number of children, typically two. As a result, the order information of siblings is of relatively low importance, making the use of children nodes a viable and efficient approach. Moreover, we are able to effectively retain information about the children by averaging their representations.

\begin{figure}[H]
    \centering
    \includegraphics[width=0.5\textwidth]{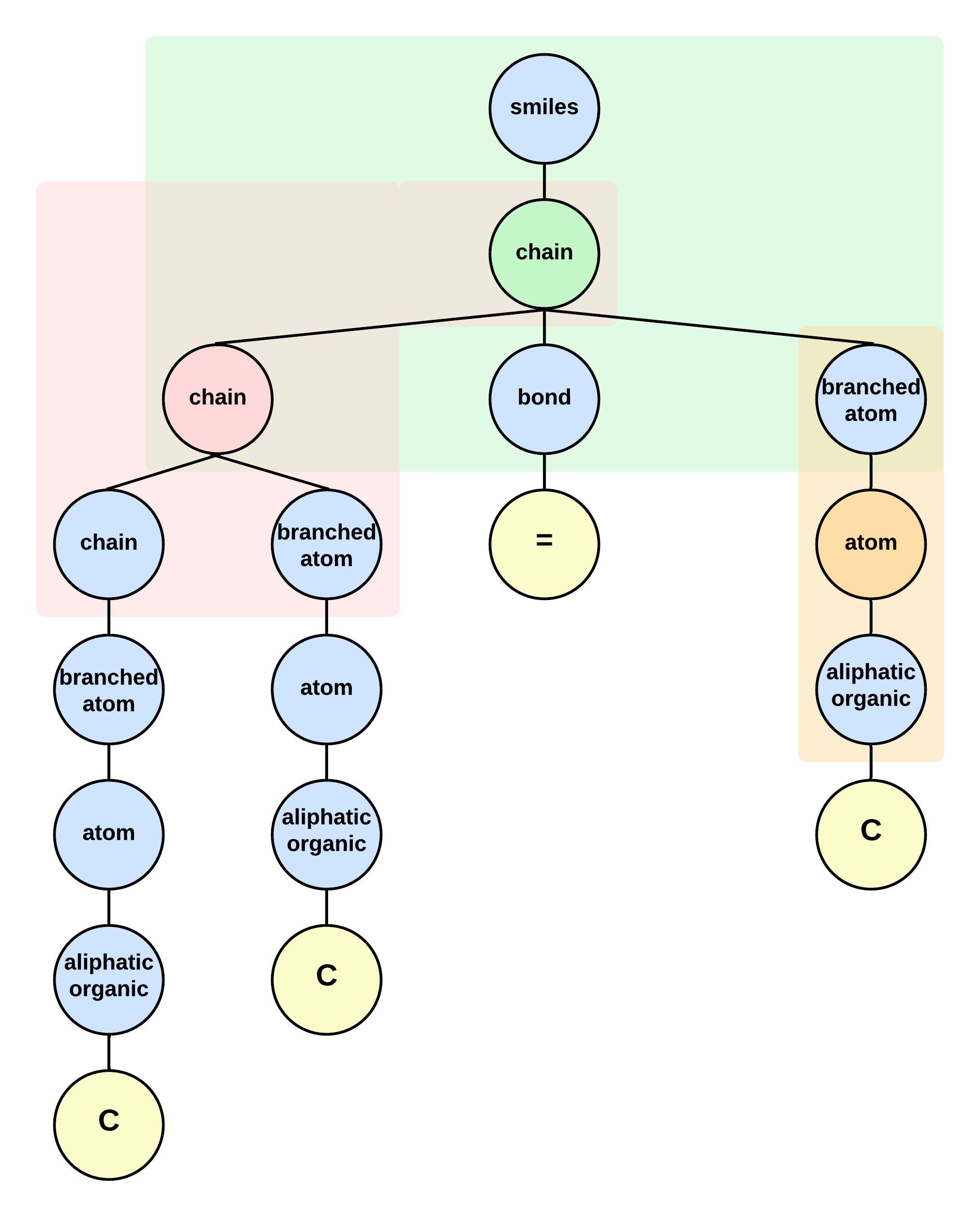}
    \caption{The grammar tree for propene \texttt{CC=C} and three example convolution windows in red, green, and orange. Each convolution window operates on the respectively colored node, thereby providing the node with information about its parent and children.}
    \label{fig:tcbtree}
\end{figure}

To visualize the behavior of the TCB, we provide an example of its operation on the grammar tree for propene in Figure \ref{fig:tcbtree}. Each colored region represents the convolution window centered around the corresponding node. It is evident from this illustration that the TCB enables the model to explicitly incorporate information about the surrounding local structures. This prior chemistry-aware knowledge allows the architecture to more effectively model the relationships between molecules, as it possesses a richer structural details.

\subsection*{Beam search}
\begin{figure}[H]
    \centering
    \includegraphics[width=\textwidth]{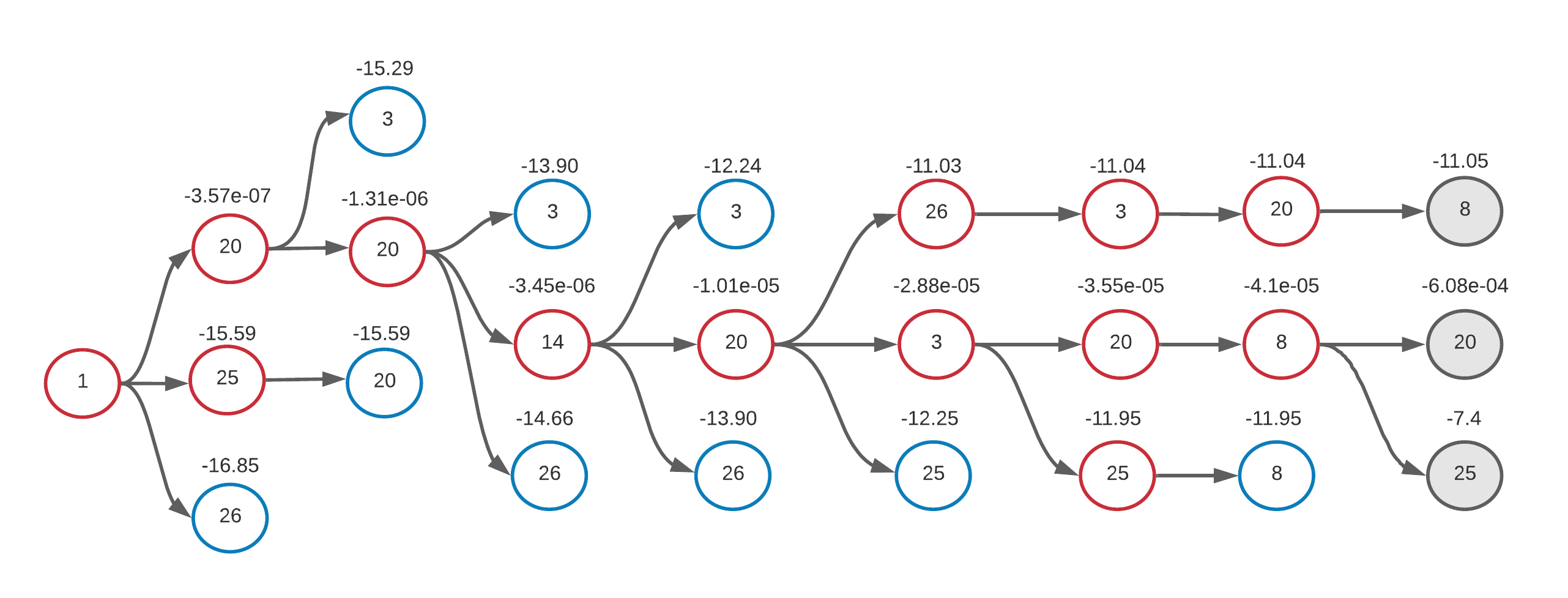}
    \caption{A schematic for a partially completed beam search procedure with beam size $B = 3$. At each decoding step, the 3 most likely sequences are preserved and used as the search frontier. The $B$ most likely tokens for each sequence is computed and the top-$B$ sequences are used for the next step. The completed sequences are then used to reconstruct the corresponding SMILES string. The log-likelihood values are indicated above each node in the schematic.}
    \label{fig:beamsearch}
\end{figure}

During inference, we evaluate the model's performance using a beam search decoding strategy to autoregressively predict the output. The decoding process begins with a \texttt{<START>} token, and the transformer generates the next token from the latent space using the current predicted sequence as input. The beam search procedure, with a specified beam size $B$, determines the top-$B$ tokens at each step based on their likelihood and retains the top-$B$ sequences as the search frontier for the next step. If $B = 1$, the transformer employs a greedy strategy, selecting the maximum likelihood token at each decoding step. By using beam search, we can thoroughly evaluate our model's performance and compare it with the top-k accuracy reported in other similar works. \textcolor{red}{We use a beam search size of $B = 10$ for all evaluation metrics outlined in Section ``\nameref{sec:results}".} A schematic of the beam-search decoding procedure with $B = 3$ is depicted in Figure \ref{fig:beamsearch}. 

\section*{Dataset and model training}\label{sec:dataset-model-training}
In this section, we present comprehensive details about the datasets used for training the retrosynthesis models, as well as information regarding the model training aspects.

\subsection*{Dataset}
For training and evaluation, we use the USPTO-50K retrosynthesis prediction dataset, which is a curated subset of the US Patents and Trademark Office's (USPTO) database \cite{lowe2012extraction}. This dataset has been further classified into ten distinct reaction classes \cite{schneider2016s}, and the number of reactions in each class is detailed in Appendix 2. The filtered dataset exclusively contains reactants and products, with the reagent information removed, and the SMILES strings are canonicalized. It is a widely used benchmark dataset in the literature for evaluating retrosynthesis model performance, offering the flexibility to train models for both known and unknown reaction class scenarios. 

Since our model uses the SMILES grammar-based tree representation, we parse the input molecules' SMILES strings from the database using the grammar outlined in Section ``\nameref{sec:SMILESgrammar}". As a result, certain molecules may not be \textit{in grammar} are thus are not included in the model training process. However, it is worth noting that expanding the grammar to include these molecules is straightforward by adding corresponding rules for additional elements such as Si, Pt, Zn, Mg, and others. Table \ref{tab:datadiff} provides a summary of the reaction database, indicating the number of reactions that are in grammar, along with the train, validation, and test-set splits.

\begin{longtable}{@{}llllr@{}}
\caption{An overview of the retrosynthesis dataset}
\label{tab:datadiff}\\
\toprule
\textbf{Dataset} & \multicolumn{1}{c}{\textbf{train}} & \multicolumn{1}{c}{\textbf{valid}} & \multicolumn{1}{c}{\textbf{test}} & \multicolumn{1}{c}{\textbf{total}} \\* \midrule
\endhead
\bottomrule
\endfoot
\endlastfoot 
\textbf{USPTO-50K (retrosynthesis)} & & & & \\
with (sanitized) molecules & &  &  & 50,037 \\
in grammar & 38,899 & 4,849 & 4,846 & \textbf{48,594} \\
 \bottomrule
\end{longtable}

\subsection*{Model training}
\begin{figure}[H]
    \centering
    \includegraphics[width=\textwidth]{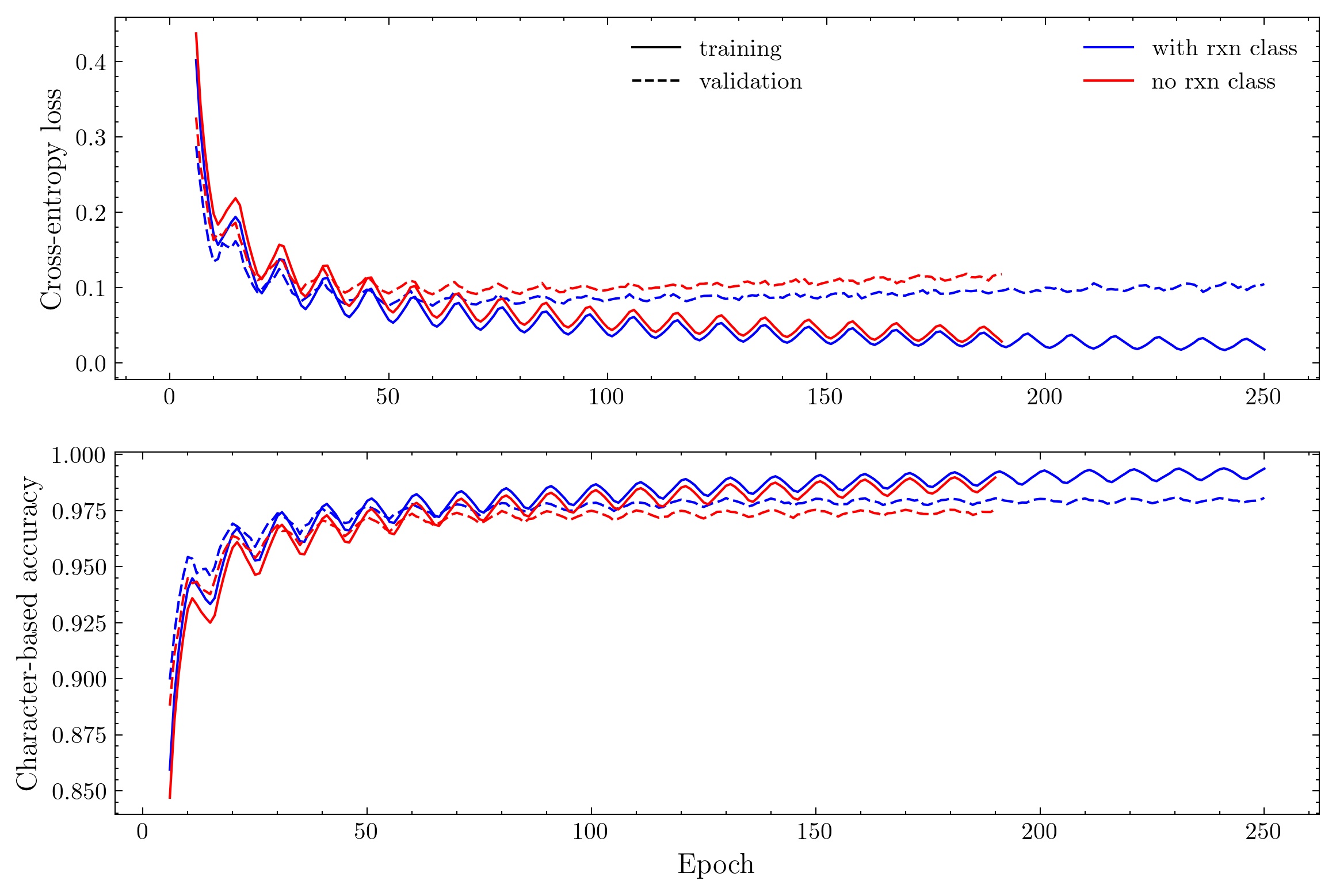}
    \caption{\textcolor{red}{The cross-entropy loss and character-based accuracy for the training and validation set. We train the with reaction class model for 250 epochs and the without reaction class model for 190 epochs. In both cases, we do not use an early stopping criteria for evaluation on the test set. The oscillatory behavior of the training loss and accuracy is due to the cyclic learning rate scheduler, which has a cycle length of $n = 10$.}}
    \label{fig:train-val-curves}
\end{figure}

The optimization objective was to minimize the loss function based on sparse categorical cross-entropy between the predicted and actual target sequences. We used the Adam optimizer \cite{kingma2014adam} with hyperparameters $\beta_1=0.9$, $\beta_2=0.98$, and $\epsilon=10^{-9}$. Additionally, we used a triangular cyclic learning rate schedule proposed in \cite{howard2018universal}, with a maximum learning rate $\eta_{max} = 5 \times 10^{-4}$ and a minimum learning rate $\eta_{min} = 1 \times 10^{-4}$. The number of epochs per cycle was set to $n = 10$ and learning rate decay per cycle was $\gamma = 0.98$. During the training stage, we incorporated dropout layers with probability $p = 0.2$ in both the attention sublayer and TCBs to prevent overfitting. We fixed the lengths of the input and output representations to the model at 350 and 121, respectively, based on the longest representation lengths observed in the training set. The retrosynthesis model for known reaction class was trained for 250 epochs, while the model for the unknown reaction class was trained for 190 epochs. These choices ensured adequate training convergence and achieved desirable model performance. \textcolor{red}{In Figure \ref{fig:train-val-curves}, we present the cross-entropy loss and character-based accuracy plotted for both the training and validation sets throughout the entire training phase. Notably, we do not employ early stopping based on validation loss since previous works \cite{karpov2019transformer} have indicated that such an approach potentially undermines model performance.}

\section*{Results}\label{sec:results}
In this section, we present the performance of the trained models on the test set. To assess retrosynthesis prediction, we first define the model evaluation metrics. Additionally, we compare our model's performance with other similar works in this field to benchmark its effectiveness. Finally, we will discuss the advantages of our framework and address its limitations.


Given that a product may have multiple correct retrosynthesis pathways, we consider not only exact matches between predicted reactions and ground truth but also the chemical similarity of precursor molecules. This evaluation is important because chemically similar molecules are likely to produce feasible and correct reactions in practice. Therefore, we incorporate similarity-based metrics to provide a more comprehensive and practical assessment of the model's performance.


To determine chemical similarity, we use the Tanimoto index, one of the most widely used and reliable metrics for computing structural similarity between molecules \cite{bajusz2015tanimoto}. The Tanimoto coefficient $T_c$ is a value ranging from $0$ to $1$, representing the fraction of common molecular fingerprints shared between two molecules. For two molecules $X$ and $Y$ with fingerprint sets $A$ and $B$ respectively, the Tanimoto coefficient is computed as: \[ T_c (X,Y) = \frac{A \cap B}{A + B - A \cup B}. \] A coefficient of zero implies that the molecules have no common fingerprints, while a coefficient of one indicates two identical molecules. Although no specific coefficient distinguishes similar and dissimilar molecules, we adopt the commonly used threshold of $T_c = 0.85$ for defining bioactive similarity. In other words, we consider two molecules bioactively similar if their Tanimoto score satisfies $0.85 \leq T_c \leq 1$.


\subsection*{Evaluation metrics}
\begin{figure}[H]
    \centering
    \includegraphics[width=0.8\textwidth]{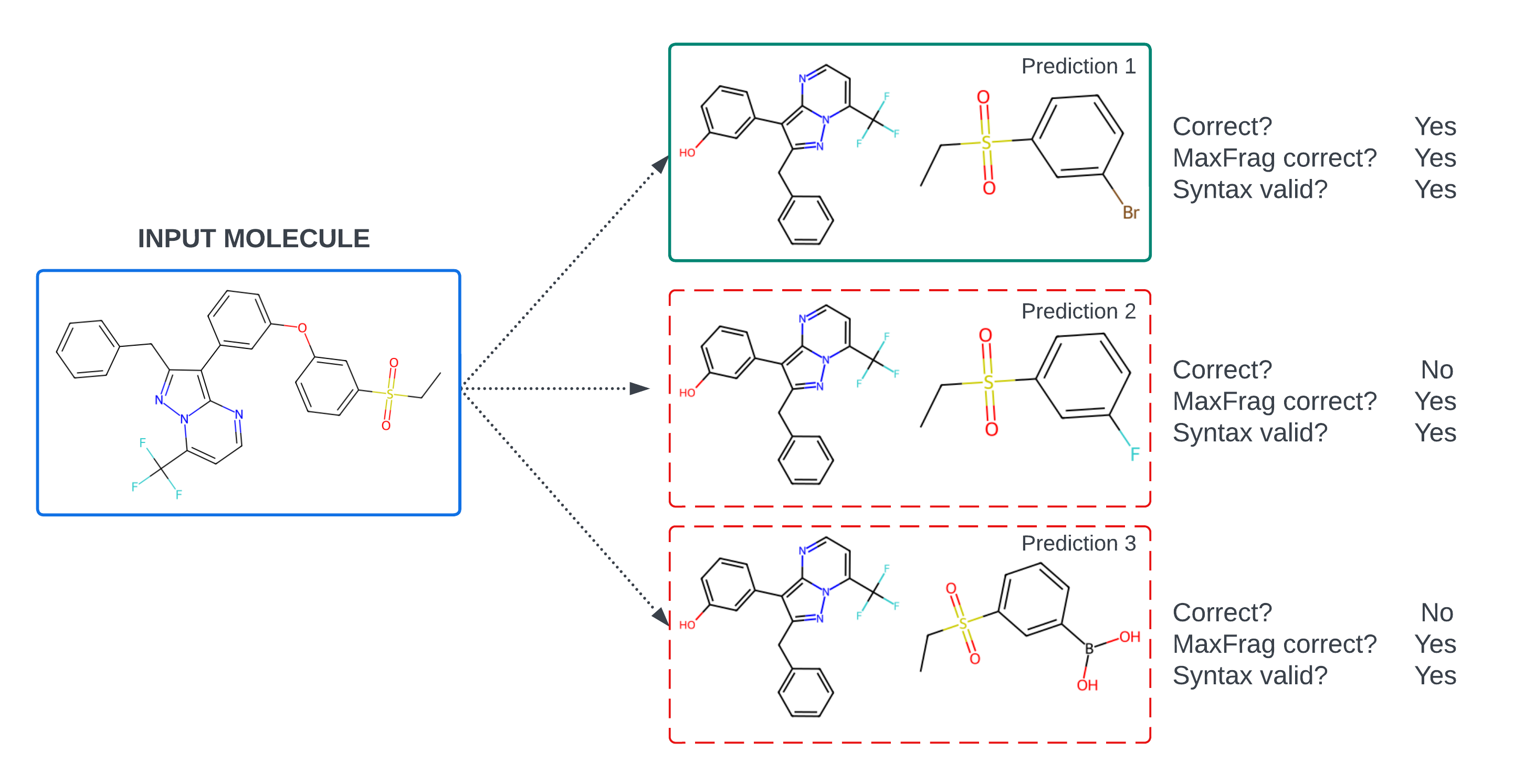}
    \caption{Top-3 predictions generated by our model for a given input molecule in the test-set. The first (most-likely) prediction matches the ground truth, the second is chemically feasible but incorrect since it doesn't exactly match the ground truth, and third prediction is chemically incorrect. All predictions are syntactically valid and has correctly predicted maximal fragment.}
    \label{fig:retro-eval-metrics}
\end{figure}

We define the following performance metrics to evaluate our models: accuracy, fractional accuracy, MaxFrag accuracy, MaxFrag bioactive similarity rate (BASR), and invalid rate. Accuracy measures the fraction of reactions in which the predicted precursor molecules perfectly match the ground truth. Fractional accuracy measures the proportion of correctly predicted precursor molecules relative to the total number of ground truth precursors. MaxFrag accuracy represents the prediction accuracy of the maximal fragment, which corresponds to the longest precursor molecule based on SMILES length. MaxFrag BASR quantifies the fraction of reactions where the predicted maximal fragment is bioactively similar (with a Tanimoto score between $0.85$ and $1$) to the ground truth maximal fragment. Lastly, the invalid rate calculates the percentage of syntactically or grammatically invalid SMILES predictions. A schematic illustrating these metrics for an example molecule in the test-set is presented in Figure \ref{fig:retro-eval-metrics}.

Based on the observations from the figure, we notice that a prediction is considered incorrect if it does not exactly match the set of precursors in the ground truth, even if it is chemically feasible. For example, the most likely prediction (prediction 1) matches the ground truth precursors and is considered correct. However, the second most likely prediction (prediction 2) contains a fluorine (F) instead of bromine (Br) for one of the precursors. Although both elements are halogens and are chemically correct, the prediction is still deemed incorrect when computing accuracy. As a result, the reported exact-match accuracy serves as a conservative lower bound on the model's performance. In practical scenarios, the actual model performance is likely to be better than expected based solely on exact-match accuracy.


\subsection*{Retrosynthesis prediction}

\begin{figure}[H]
    \centering
    \begin{subfigure}[b]{0.8\textwidth}
         \centering
         \includegraphics[width=\textwidth]{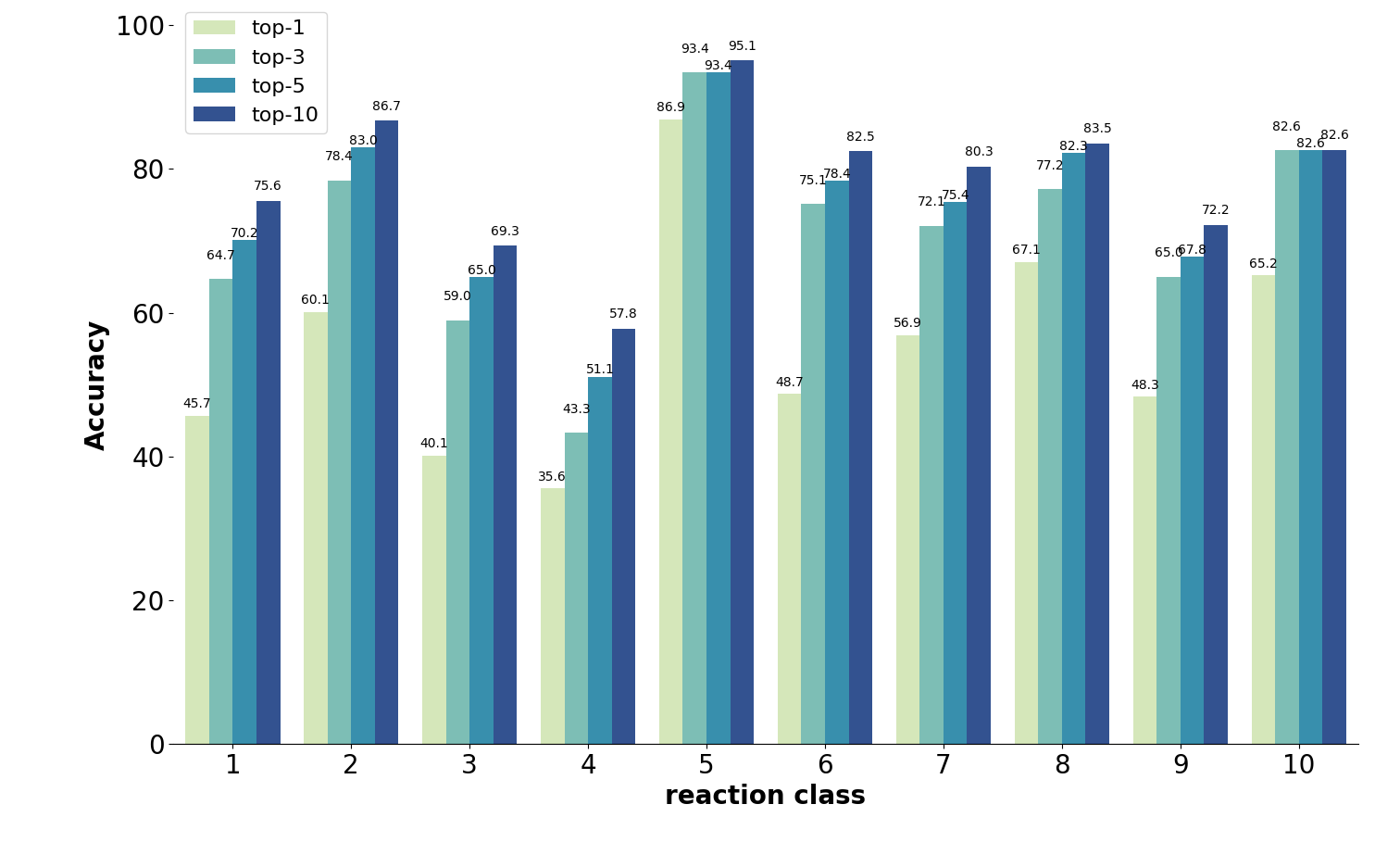}
         \caption{}
         \label{fig:acc_withclass}
     \end{subfigure}
     \begin{subfigure}[b]{0.8\textwidth}
         \centering
         \includegraphics[width=\textwidth]{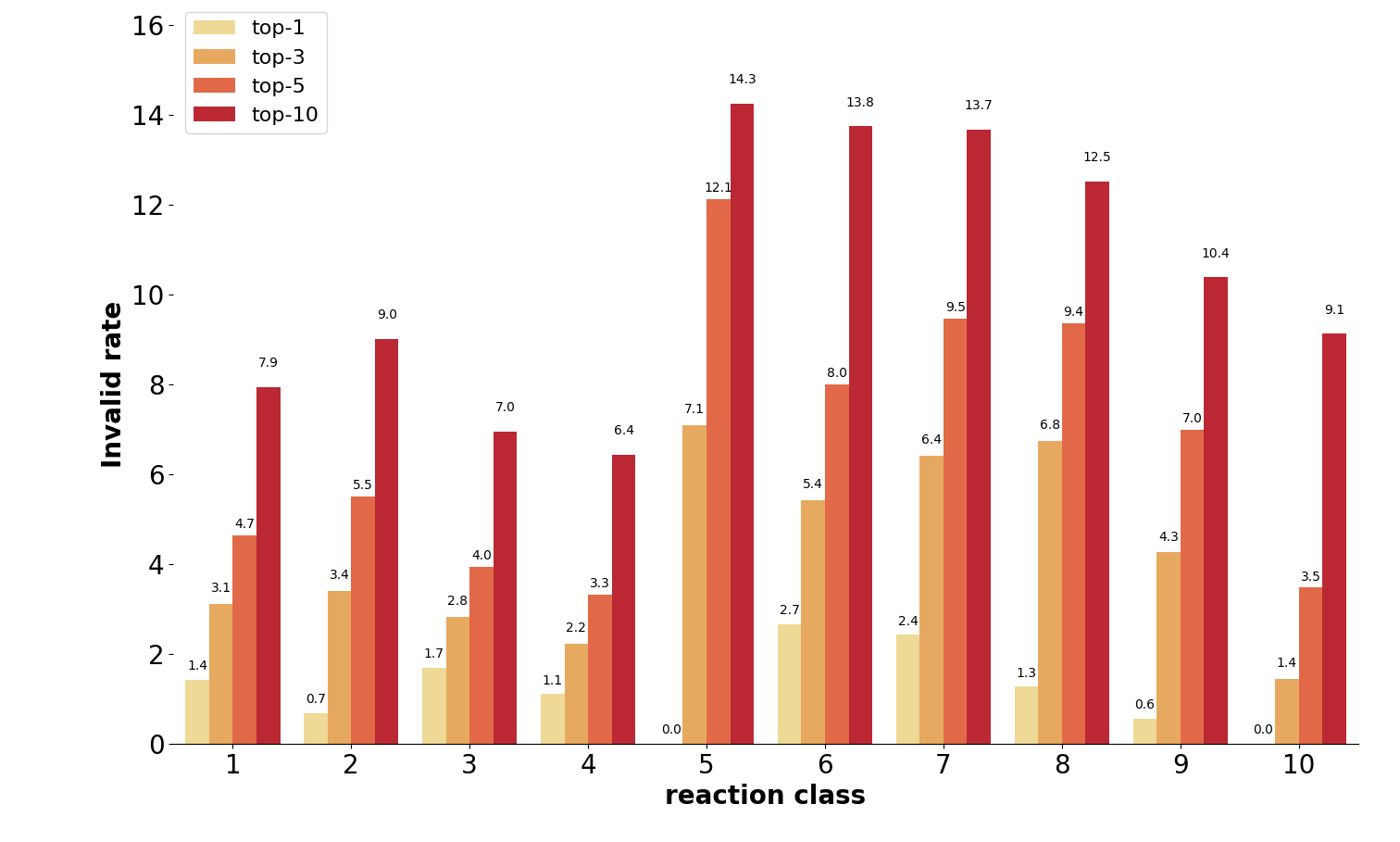}
         \caption{}
         \label{fig:inv_withclass}
     \end{subfigure}
    \caption{The class-wise top-$1$, top-$3$, top-$5$, and top-$10$ prediction accuracy (a) and invalid SMILES string rate (b) for retrosynthesis prediction with known reaction class.}
    \label{fig:rxnclass_withclass}
\end{figure}

\begin{figure}[H]
    \centering
    \begin{subfigure}[b]{0.8\textwidth}
         \centering
         \includegraphics[width=\textwidth]{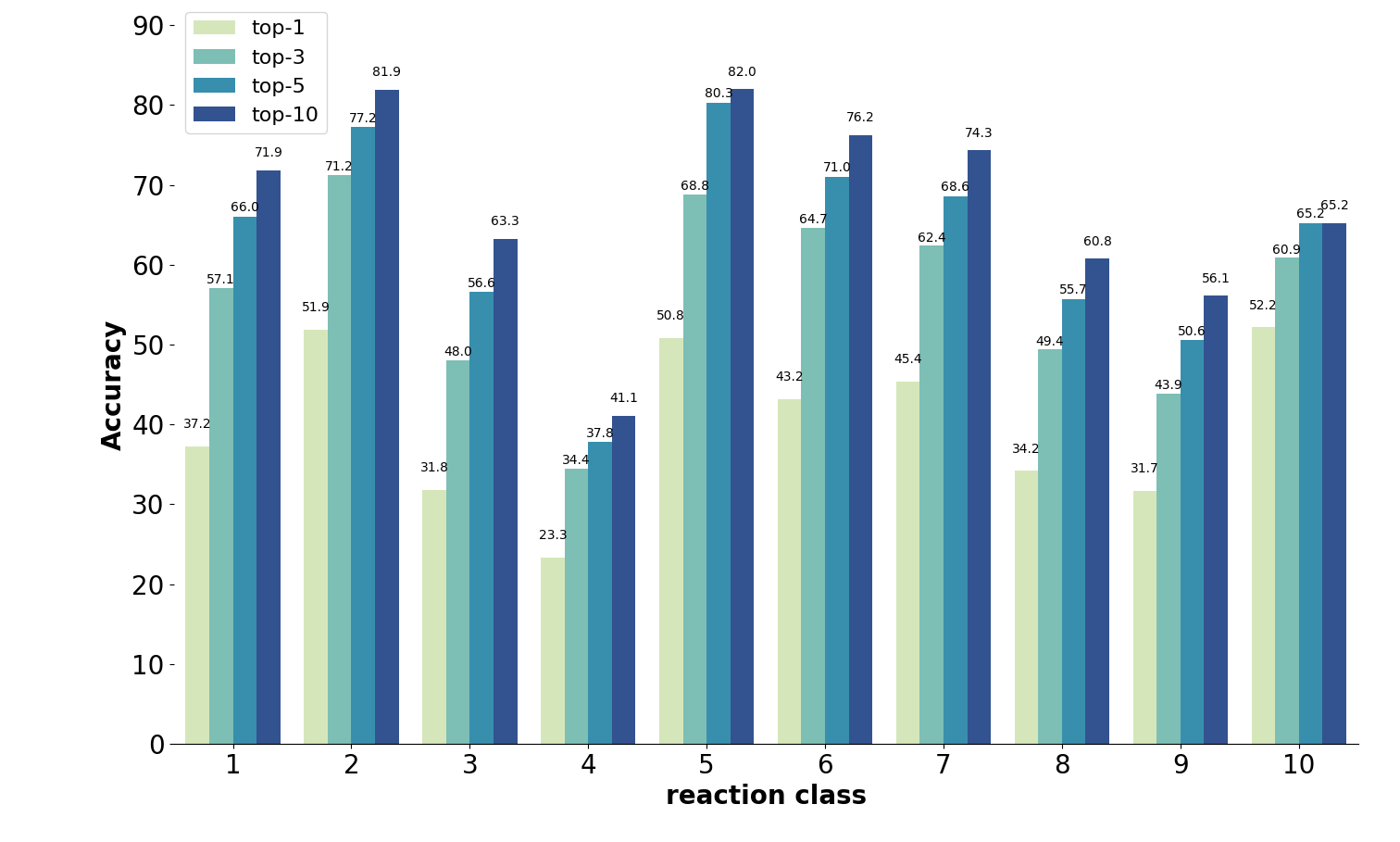}
         \caption{}
         \label{fig:acc_noclass}
     \end{subfigure}
     \begin{subfigure}[b]{0.8\textwidth}
         \centering
         \includegraphics[width=\textwidth]{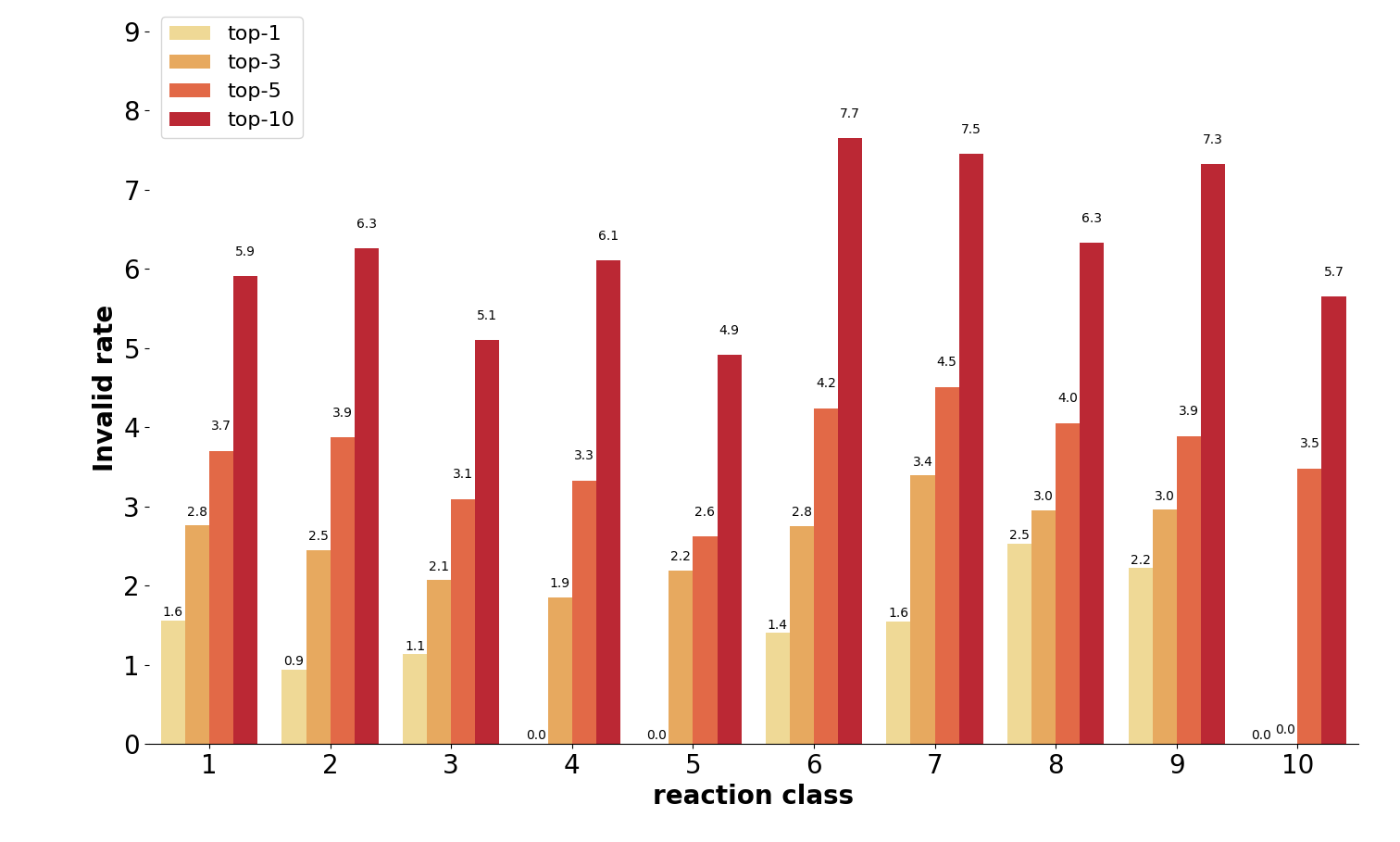}
         \caption{}
         \label{fig:inv_noclass}
     \end{subfigure}
    \caption{The class-wise top-$1$, top-$3$, top-$5$, and top-$10$ prediction accuracy (a) and invalid SMILES string rate (b) for retrosynthesis prediction with unknown reaction class.}
    \label{fig:rxnclass_noclass}
\end{figure}

\paragraph{\textbf{Performance evaluation metrics}}
Table \ref{tab:retrometrics1} presents the performance evaluation measures calculated on the test set of the USPTO-50K dataset for both the known and unknown reaction class scenarios. Additionally, we compute the class-wise prediction accuracy and invalid rate for both known and unknown reaction class scenarios in Figure \ref{fig:rxnclass_withclass} and Figure \ref{fig:rxnclass_noclass}, respectively. This granular evaluation provides insights into how the model performs in distinct reaction categories and in different reaction contexts. \textcolor{red}{Note that the retrosynthesis prediction accuracies for certain reaction classes -- reaction class 3 (corresponding to C--C bond formation) and class 4 (corresponding to heterocycle formation) are lower compared to other reaction classes. This could possibly be attributed to the inherent complexity of such reactions in terms of dependence on key reagents that determine reaction outcomes (and hence retrosynthesis) and a wide array of reactions (for reaction class 3) that are grouped together in the same reaction class -- such as aldol reactions,  Diels–Alder reaction, Grignard reaction, cross-coupling reactions, the Michael reaction,  the Wittig reaction, and so on. The combined effect of these two factors could possibly be the reason behind relatively lower accuracy for reaction class 3. Similarly, for reaction class 4, lower prediction accuracy could be attributed to a lack of enough training examples. The complete list of reaction classes and corresponding number of examples in the train, validation, and test set are listed in Appendix 2 in Table \ref{tab:distributionreactionclass} and corresponding class-wise
performance metrics in the supplementary information.}

\begin{table}[H]
\caption{Retroynthesis models' performance metrics on the test set}
\label{tab:retrometrics1}
\centering
\begin{tabular}{@{}llllll@{}}
\toprule
\multicolumn{1}{c}{\multirow{2}{*}{Performance measure}} & \multicolumn{4}{c}{top-k measure (\%)} &    \\ \cmidrule(l){2-6} 
\multicolumn{1}{c}{} & 1 & 2 & 3 & 5 & 10 \\ \midrule
\textbf{Known reaction class } & & & & & \\
Accuracy & 51.0 & 64.3 & 70.0 & 74.6 &  79.1 \\
Fractional accuracy & 64.7 & 74.9 & 79.4 & 83.2 & 86.8 \\
MaxFrag accuracy & 60.6 & 71.6 & 76.5 & 80.4 & 84.1 \\
MaxFrag BASR & 74.8 & 82.2 & 85.4 & 88.3 & 92.3 \\
Invalid rate & 1.5 & 2.7 & 4.0 & 6.0 & 9.8 \\\\

\textbf{Unknown reaction class } & & & & & \\
Accuracy & 41.6 & 54.0 & 60.4 & 67.6 & 73.1 \\
Fractional accuracy & 52.0 & 63.7 & 69.8 & 76.3 & 81.3 \\
MaxFrag accuracy & 49.0 & 60.5 & 66.5 & 72.7 & 77.8 \\
MaxFrag BASR & 60.0 & 70.8 & 76.4 & 82.2 & 86.8 \\
Invalid rate & 1.3 & 2.0 & 2.6 & 3.8 & 6. 
                     \\ \bottomrule
\end{tabular}%
\end{table}


\clearpage

\clearpage


\paragraph{\textbf{Comparison with other works}} 
In this section, we benchmark the performance of our G-MATT model against other transformer-based approaches reported in the literature. \textcolor{red}{All models in the comparison are trained and evaluated on the same USPTO-50K dataset used for training and evaluating our model.} To provide a comprehensive comparison, we evaluate the top-k prediction accuracy and top-k invalid rate as shown in Table \ref{tab:retro-comparison}. Our findings demonstrate that for both the known and unknown reaction class scenarios, the proposed G-MATT model achieves nearly state-of-the-art performance in terms of prediction accuracy and invalid rate.

\begin{table}[H]
\caption{Comparison with other transformer-based retrosynthesis models trained on USPTO-50K dataset}
\label{tab:retro-comparison}
\centering
\begin{tabular}{@{}lllllllll@{}}
\toprule
\multicolumn{1}{c}{\multirow{2}{*}{model}} & \multicolumn{4}{c}{top-k accuracy (\%)} & \multicolumn{4}{c}{top-k invalid rate (\%)} \\ \cmidrule(l){2-9} 
\multicolumn{1}{c}{}   & 1 & 3 & 5 & 10 & 1 & 3 & 5 & 10 \\ \midrule
\textbf{Known reaction class}   &   &   &   &    &   &   &   &    \\
Liu et al. \cite{liu2017retrosynthetic}  &  37.4 & 52.4  & 57.0  & 61.7   & 12.2  & 15.3  & 18.4  &  22.0  \\
Grammar seq2seq \cite{mann2021retrosynthesis}  &  43.8 & 57.2  & 61.4  & 66.6   & 4.4  & 7.2  & 8.4  &  9.6  \\
SCROP \cite{scrop2020}           & 59.0  & 74.8  & 78.1  &  81.1  &   &   &   &   \\
Lin et al. \cite{lin2020automatic}     & 54.3  & 74.1  &  79.2 & 84.4   &  2.3 & 4.9  & 7.0  &  12.1  \\
T2T \cite{duan2020retrosynthesis}  & 54.1  & 68.0  &  69.0 &  70.1 &   &   &   & \\
\textbf{G-MATT (proposed) }     & 51.0  &  70.0 & 74.6  &  79.1 & 1.5  & 4.0 & 6.0 & 9.8   \\\\
\textbf{Unknown reaction class }&   &   &   &    &   &   &   &    \\
Liu et al. \cite{liu2017retrosynthetic}  &  28.3 & 42.8  & 47.3  & 52.8   &   &   &   &   \\
Grammar seq2seq \cite{mann2021retrosynthesis}  & 32.1  & 44.3  &  48.9 & 54.0   & 5.1  & 7.4  &  8.4 & 9.7   \\
Chen et al. \cite{chen2019learning}   & 39.1  & 62.5  &  69.1 & 74.5    &   &   &   &    \\
Karpov et al. \cite{karpov2019transformer}  & 40.6  & 42.7  & 63.9  & 69.8   &   &   &   &    \\
SCROP \cite{scrop2020}           & 43.7  & 60.0  & 65.2  &  68.7  &  0.7 & 1.4  & 1.8  & 2.3   \\
Lin et al. \cite{lin2020automatic}                  & 42.0  & 64.0  &  71.3 & 77.6   & 2.2  & 3.7  & 4.8  & 7.8   \\
Tied two-way TF  \cite{kim2021valid}  & 47.1  & 67.1  & 73.1  &  76.3  &  0.1 & 0.2  & 0.6  & 10.2   \\
\textbf{G-MATT (proposed)}      & 41.6  & 60.4  & 67.6  &  73.1 & 1.3 & 2.6 & 3.8 & 6.3      \\ \bottomrule
\end{tabular}%
\end{table}


In the case of the known reaction class scenario, G-MATT's top-k prediction accuracy is, on average, only 4.3\% lower than that of Lin et al.'s state-of-the-art model \cite{lin2020automatic}. However, G-MATT has an 1.25\% improvement in invalid rate over Lin et al.'s model. Similarly, for the unknown reaction class scenario, G-MATT's top-k prediction accuracy is within 3.1\% of Lin et al.'s model, while outperforming it by 1.13\% on the invalid rate. Additionally, when compared to SCROP, which utilizes an additional transformer model to correct incorrect predictions made by a primary retrosynthesis transformer model, G-MATT, with its single transformer architecture, outperforms the prediction accuracy for the unknown reaction class scenario across various top-k values.


It is worth noting that G-MATT, despite its relative simplicity compared to other approaches in the literature, achieves nearly state-of-the-art performance in terms of prediction accuracy and invalid prediction rate. This can be attributed to several key characteristics that distinguish G-MATT. Firstly, G-MATT benefits from richer input representations based on SMILES grammar trees, as opposed to solely relying on linear SMILES strings. Secondly, the inclusion of tree positional encodings allows the model to accurately capture the hierarchical structure present in the trees. Lastly, the use of tree convolution blocks within the transformer architecture enables the model to attend to local molecular structures, such as functional groups, thereby providing valuable contextual information. These factors collectively contribute to G-MATT's impressive performance.

While G-MATT already demonstrates competitive results, we acknowledge that further improvements are possible by incorporating additional model training strategies from prior works in the literature. Our primary objective is not to pursue the state-of-the-art performance, but rather to showcase the value of utilizing alternative SMILES grammar-based tree representations for reaction prediction tasks. The effectiveness of G-MATT highlights the potential benefits of considering underlying molecular structures and hierarchical information, leading to improved performance in retrosynthesis prediction.


\paragraph{\textbf{Near-miss predictions}} 
To further analyze the model's performance, we conduct a analysis of incorrect predictions from a chemistry standpoint. Specifically, we compute the Tanimoto similarity of incorrectly predicted maximal fragment molecules with the ground truth. We focus on the maximal fragment since there are multiple molecules in the target reaction pathway. This analysis provides us with insights into the degree of similarity between the incorrectly predicted precursors and the ground truth, which is important as chemically similar precursors are likely to lead to feasible and accurate reactions in practice. Table \ref{tab:tanimoto-sim} below presents the percentile splits of incorrect predictions, categorized based on their Tanimoto similarities.

\begin{table}[H]
    \centering
    \begin{tabular}{@{}lccc@{}}
    \toprule
    \multicolumn{1}{c}{\multirow{1}{*}{Model}} & \multicolumn{3}{c}{Tanimoto similarity (\%) (incorrect predictions only)}\\ \cmidrule{2-4} 
    \multicolumn{1}{c}{} & $0.5 \leq T_c$ & $0.7 \leq T_c$ & $0.85 \leq T_c$ \\ \midrule
    Known reaction class & 66.6 & 46.8 & 31.5 \\
    & & & \\
    Unknown reaction class & 78.8 & 63.7 & 48.7 \\
    \bottomrule
    \end{tabular}
    \caption{Distribution of Tanimoto coefficient scores across \textit{incorrect} top-1 predictions for both the models}     \label{tab:tanimoto-sim}
\end{table}



\begin{figure}[H]
    \centering
    \begin{subfigure}[b]{0.8\textwidth}
        \centering
        \includegraphics[width=0.8\textwidth]{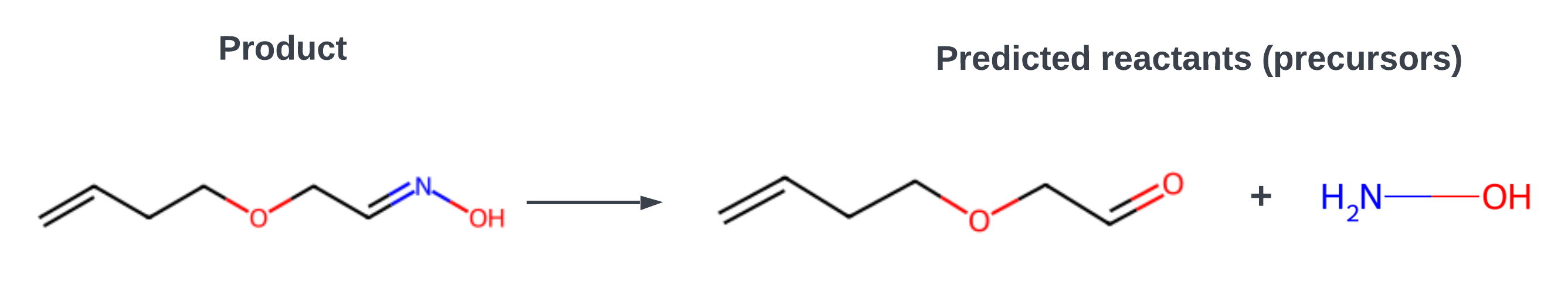}
        \caption{An example top-1 prediction to study the transformer attention map}
        \label{fig:attention-rkn}
     \end{subfigure}\\
     \begin{subfigure}[b]{0.9\textwidth}
     \centering
        \includegraphics[width=\textwidth]{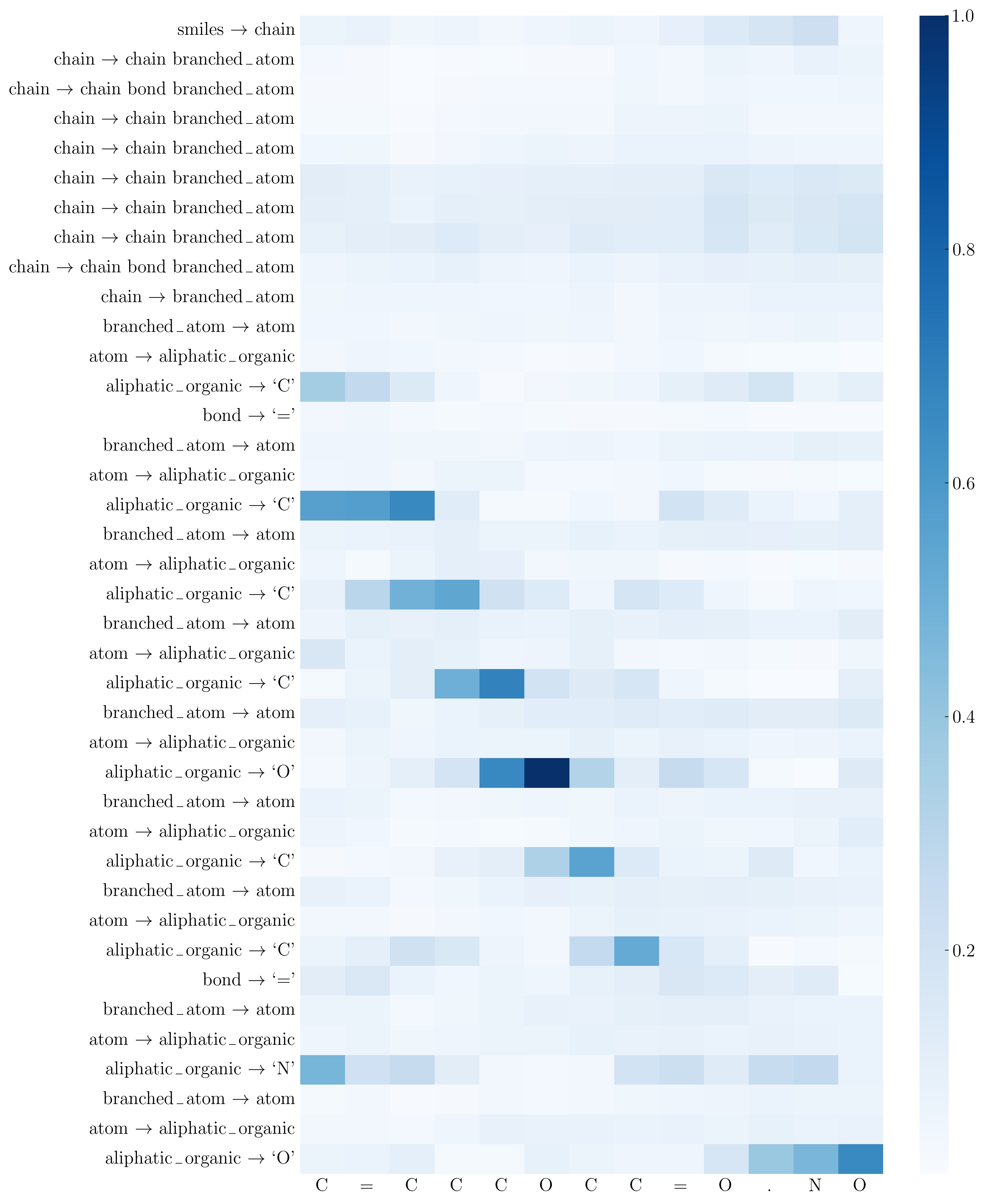}
        \caption{Average attention scores for top-1 prediction on the retrosynthesis reaction \texttt{C=CCCOCC=NO} $\to$ \texttt{C=CCCOCC=O.NO} extracted from the transformer decoder cross-attention sublayer}
        \label{fig:heatmap}
     \end{subfigure}
    \caption{Molecular attention map for the SMILES grammar tree of the product and SMILES strings of the reactants (precursors) for an example reaction}
    \label{fig:attn_heatmap}
\end{figure}

\paragraph{\textbf{Attention-map for molecules}}
To gain deeper insights into the grammar-based molecular representation, we conduct an analysis of the attention weights computed by the attention sublayers. We specifically focus on the cross-attention modules in the decoder and calculate the average attention scores across all of them. The resulting attention heatmap is showcased in Figure \ref{fig:attn_heatmap}. Higher attention scores indicate a more pronounced correlation between the corresponding tokens, signifying their greater relevance in the translation process. This investigation demonstrates how the transformer utilizes the grammar-based representation to identify molecular structures.

The attention mechanism exhibits high attention scores along the diagonal, indicating its ability to correctly associate atoms and bonds in the input molecule with those in the precursor molecules during translation. For instance, Figure \ref{fig:attn_heatmap} illustrates an example reaction where the bond disconnection site is the \texttt{N} atom in the \texttt{-N=O} functional group. From the attention map, we observe that both \texttt{aliphatic\_organic $\to$ O} and \texttt{aliphatic\_organic $\to$ N} have the highest attention scores for the \texttt{`.'} token, which serves as the separator between the precursor molecules. This confirms that the model has accurately identified the bond disconnection site. This pattern is further validated through additional examples provided in Appendix 3 where the bond disconnection sites -- \texttt{aliphatic\_organic $\to$ S} in Figure \ref{fig:attn_heatmap2} and \texttt{aromatic\_organic $\to$ c} in Figure \ref{fig:attn_heatmap3} -- have the highest attention scores with molecule separate token. Furthermore, we observe that atoms in the precursor molecules (horizontal axis) often have high attention scores with multiple adjacent atoms or bonds in the corresponding product (vertical axis), and vice versa. These findings strongly suggest that due to the convolutional operations on the hierarchical tree representation, G-MATT considers the context of neighboring atoms while making predictions and identifies the importance of local structures within the molecules.  \textcolor{red}{This capability captures more comprehensive relationships between atoms and bonds, improving the accuracy of predicted pathways.}

Another noteworthy observation is that G-MATT's attention maps are generally sparse, with several input tokens having relatively lower cross-attention scores. This sparsity indicates the efficiency of the tree representation, which inherently imposes structure on the tokens. Consequently, the model captures both the semantics and syntactical structure of tokens, allowing it to focus on the most crucial tokens while resolving uncertainty during predictions. The model can infer the remaining tokens from the structural constraints it has learned. \textcolor{red}{This ability to attend selectively to important tokens contributes to the model's performance despite its relatively low complexity.} Additional cross-attention maps for the G-MATT model are provided in Appendix 3.

\section*{Data Availability and Reproducibility Statement}
The dataset used in this work is the standard and publicly available USPTO-50K retrosynthesis prediction dataset extracted from US Patents and Trademark Office's (USPTO) database \cite{lowe2012extraction}. The numerical data from Figures \ref{fig:rxnclass_withclass} and \ref{fig:rxnclass_noclass} are tabulated in the Supplementary Material. Numerical data for the attention maps in Figures \ref{fig:attn_heatmap}, \ref{fig:attn_heatmap2}, and \ref{fig:attn_heatmap3} are provided as .zip files in the Supplementary Information. For generating the parse-trees and extracting grammar-based features, we used the Natural Language ToolKit (NLTK) $3.4.5$ library. All the models were trained using TensorFlow version $2.1$ and implemented in Python version $3.7$. Additionally, the molecular datasets were processed using the $2019$ release of the RDKit library.


\section*{Conclusions}\label{sec:conclusions}
In this work, we introduced a novel tree-based transformer architecture for retrosynthesis, which represents molecular inputs as hierarchical trees instead of character-based SMILES string representations. \textcolor{red}{Our tree transformer significantly improves upon previous approaches in three key aspects. Firstly, we leverage the SMILES grammar to represent molecules, incorporating explicit chemistry information to make the model more domain-aware and robust. Secondly, we introduce tree positional encodings to indicate the position of nodes in the grammar tree, thereby fully utilizing the tree hierarchy structure and inherent structural relationships. Lastly, we employ tree convolutions, providing nodes with progressively more contextual information. This context is important in identifying molecular characteristics, such as functional groups or other local structures, which might be ambiguous the traditional SMILES string representation. Throughout our experiments, we demonstrated that explicitly incorporating the tree structure in the transformer model takes full advantage of the benefits provided by a grammar-based tree representation. This results in enhanced performance and a deeper understanding of the underlying chemistry, making our tree-based transformer a promising approach for retrosynthesis prediction.}

In the known reaction class scenario, we achieved a top-1 prediction accuracy of $51.0\%$ (top-10 accuracy of $79.1\%$), fractional accuracy of $64.7\%$, and a syntactic invalid rate of $1.5\%$. Our model performed similarly for the unknown reaction class case, achieving a top-1 prediction accuracy of $41.6\%$ (top-10 accuracy of $73.1\%$), fractional accuracy of $52.0\%$, and a syntactic invalid rate of $1.3\%$. The attention score visualizations provided insights into the model's robustness, as it accurately identified reaction centers, considered the surrounding context of atoms and bonds, and effectively learned structural constraints. Remarkably, the G-MATT framework, despite its relative simplicity, attained nearly state-of-the-art performance in terms of prediction accuracy and invalid rate. 

\textcolor{red}{Going forward, our focus for future work would be to incorporate additional reaction conditions in the retrosynthesis prediction framework, address issues related to underrepresented reaction classes during model training, and further bifurcate the reaction classes into more specific reaction types would help in improving the model performance further.} With additional model training strategies, such as model weight averaging, customized learning schedules, and exhaustive hyperparameter search, we envision further improvements in G-MATT's performance. \textcolor{red}{ Moreover, performing multi-step retrosynthesis by combining our single-step model with techniques such as Monte Carlo tree search would be our future direction of research.}

\section*{CRediT authorship contribution statement}
\textbf{Kevin Zhang}: Conceptualization, Formal analysis, Methodology, Writing – original draft, Writing – review \& editing. \\
\indent \textbf{Vipul Mann}: Conceptualization, Formal analysis, Methodology, Writing – original draft, Writing – review \& editing. \\
\indent \textbf{Venkat Venkatasubramanian}: Conceptualization, Writing – review \& editing, Supervision, Funding acquisition

\section*{Acknowledgements}
This work was supported by the U.S. National Science Foundation (NSF) under Grant No. 2132142 and carried out at Columbia University.

\newpage
\section*{Appendix 1: SMILES grammar}
The SMILES grammar used in this work is the same as that used in our previous works \cite{mann2021predicting, mann2021retrosynthesis}. This grammar comprises 80 production rules with 24 non-terminals symbols specifying the different structural components of a SMILES string. All the production rules for the grammar used in our work are summarized in Table \ref{tab:smilesgrammar}. The first and the last production rules,  $\texttt{SMILES} \longrightarrow \texttt{CHAIN}$ and $\texttt{NOTHING} \longrightarrow \texttt{NONE}$, are additional rules included signifying the start and end of a SMILES string, which is analogous to the $\texttt{<START>}$ and $\texttt{<END>}$ tokens in natural language processing marking the beginning and the end of sentences, respectively. 

\small
\begin{longtable}{@{}ll@{}}
\caption{SMILES grammar used for retrosynthesis and forward reaction prediction. `$\mid$' separates multiple production rules applicable for the same non-terminal symbol.}
\label{tab:smilesgrammar} \\
\toprule
\textbf{S.No} & \multicolumn{1}{c}{\textbf{Production rules}} \\* \midrule
\endhead
\bottomrule
\endfoot
\endlastfoot
1 & $\texttt{SMILES} \longrightarrow \texttt{CHAIN}$   \\ 
2 & $\texttt{ATOM} \longrightarrow \texttt{BRACKET\_ATOM} ~|~   \texttt{ALIPHATIC\_ORGANIC} ~|~ \texttt{AROMATIC\_ORGANIC}$ \\ 
3 & $\texttt{ALIPHATIC\_ORGANIC} \longrightarrow \texttt{B} ~|~   \texttt{C} ~|~ \texttt{N} ~|~ \texttt{O} ~|~ \texttt{S} ~|~ \texttt{P} ~|~ \texttt{F} ~|~ \texttt{I} ~|~ \texttt{Cl} ~|~ \texttt{Br} $  \\ 
4 & $\texttt{AROMATIC\_ORGANIC} \longrightarrow  \texttt{c} ~|~ \texttt{n} ~|~ \texttt{o} ~|~ \texttt{s} ~|~ \texttt{p}$ \\ 
5 & $\texttt{BRACKET\_ATOM} \longrightarrow  \texttt{LEFT\_BRACKET}~\texttt{BAI}~\texttt{RIGHT\_BRACKET} $  \\ 
6 & $\texttt{BAI} \longrightarrow \texttt{SYMBOL} ~ \texttt{BAC} $\\
7 & $\texttt{BAC} \longrightarrow \texttt{CHIRAL} ~ \texttt{BAH} ~|~ \texttt{BAH} ~|~ \texttt{CHIRAL}$ \\
8 & $\texttt{BAH} \longrightarrow \texttt{HCOUNT} ~\texttt{BACH} ~|~ \texttt{BACH} ~|~ \texttt{HCOUNT}$ \\
9 & $\texttt{BACH} \longrightarrow \texttt{CHARGE} $ \\
10 & $\texttt{SYMBOL} \longrightarrow \texttt{ALIPHATIC\_ORGANIC} ~|~ \texttt{AROMATIC\_ORGANIC} $ \\
11 & $\texttt{DIGIT} \longrightarrow \texttt{1} ~|~ \texttt{2} ~|~ \texttt{3} ~|~ \texttt{4} ~|~ \texttt{5} ~|~ \texttt{6} ~|~ \texttt{7} $ \\
12 & $\texttt{CHIRAL} \longrightarrow \texttt{@} ~|~ \texttt{@@}$ \\
13 & $\texttt{HCOUNT} \longrightarrow \texttt{H} ~|~ \texttt{H}~ \texttt{DIGIT}$ \\
14 & $\texttt{CHARGE} \longrightarrow \texttt{-} ~|~  \texttt{+} $ \\
15 & $\texttt{BOND} \longrightarrow \texttt{-} ~|~ \texttt{=} ~|~ \texttt{\#} ~|~ \texttt{/} ~|~ \texttt{\textbackslash}$ \\
16 & $\texttt{RINGBOND} \longrightarrow \texttt{DIGIT} ~|~ \texttt{BOND}~\texttt{DIGIT} $ \\
17 & $\texttt{BRANCHED\_ATOM} \longrightarrow \texttt{ATOM} ~|~ \texttt{ATOM}~\texttt{RB} ~|~  \texttt{ATOM}~\texttt{BB} ~|~ \texttt{ATOM}~\texttt{RB}~\texttt{BB} $ \\
18 & $\texttt{RB} \longrightarrow \texttt{RB}~\texttt{RINGBOND} ~|~ \texttt{RINGBOND} $ \\
19 & $\texttt{BB} \longrightarrow \texttt{BB}~\texttt{BRANCH} ~|~ \texttt{BRANCH} $ \\
20 & $\texttt{BRANCH} \longrightarrow \texttt{LEFT\_BRACKET}~\texttt{CHAIN}~\texttt{RIGHT\_BRACKET}~|~\texttt{LEFT\_BRACKET}~\texttt{BOND}~\texttt{CHAIN}~\texttt{RIGHT\_BRACKET} $ \\
21 & $\texttt{CHAIN} \longrightarrow \texttt{BRANCHED\_ATOM} ~|~ \texttt{CHAIN}~\texttt{BRANCHED\_ATOM} ~|~ \texttt{CHAIN}~\texttt{BOND}~\texttt{BRANCHED\_ATOM} $ \\
22 & $\texttt{LEFT\_BRACKET} \longrightarrow \texttt{BRANCHED\_ATOM} ~|~ \texttt{CHAIN}~\texttt{BRANCHED\_ATOM} ~|~ \texttt{CHAIN}~\texttt{BOND}~\texttt{BRANCHED\_ATOM} $ \\
23 & $\texttt{RIGHT\_BRACKET} \longrightarrow \texttt{BRANCHED\_ATOM} ~|~ \texttt{CHAIN}~\texttt{BRANCHED\_ATOM} ~|~ \texttt{CHAIN}~\texttt{BOND}~\texttt{BRANCHED\_ATOM} $ \\
24 & $\texttt{NOTHING} \longrightarrow \texttt{NONE} $ \\
\bottomrule
\end{longtable}

\section*{Appendix 2: USPTO-50K class-wise  distribution }
Table \ref{tab:distributionreactionclass} summarizes the distribution of the various reactions across the 10 reaction classes for the USPTO-50K dataset (retrosynthesis).

\begin{longtable}{@{}lllllr@{}}
\caption{Distribution of reactions across different reaction classes that are in-grammar}
\label{tab:distributionreactionclass}\\
\toprule \small
Rxn class &  \multicolumn{1}{c}{Rxn name} & \multicolumn{1}{c}{train} & \multicolumn{1}{c}{valid} & \multicolumn{1}{c}{test} & \multicolumn{1}{r}{\textbf{total}} \\* \midrule
\endhead
\bottomrule
\endfoot
\endlastfoot 
1 &  Heteroatom alkylation and arylation  & 11,886 &1,476 & 1,478 & 14,840 \\
2 &  Acylation and related processes  & 9,358 &1,165 & 1,169 & 11,698 \\
3 &  C -- C bond formation  & 4,324 & 544& 539 & 5,407  \\
4 &  Heterocycle formation  & 710 & 89& 90 & 889\\
5 &  Protections  & 513 & 64& 62 & 639\\
6 &  Deprotections  & 6,357 &796 & 789 & 7,942 \\
7 &  Reductions  & 3,607 & 448& 452 & 4,507\\
8 &  Oxidations  & 629 &80 & 79 & 788 \\
9 &  Functional group interconversion (FGI)  & 1,434 &176 & 180 & 1,790 \\
10 & Functional group addition (FGA)             & 177 &23 & 23 &  223\\* \bottomrule
\end{longtable}

\section*{Appendix 3: Additional cross-attention maps for the G-MATT model}
\begin{figure}[H]
    \centering
    \begin{subfigure}[b]{0.8\textwidth}
        \centering
        \includegraphics[width=0.8\textwidth]{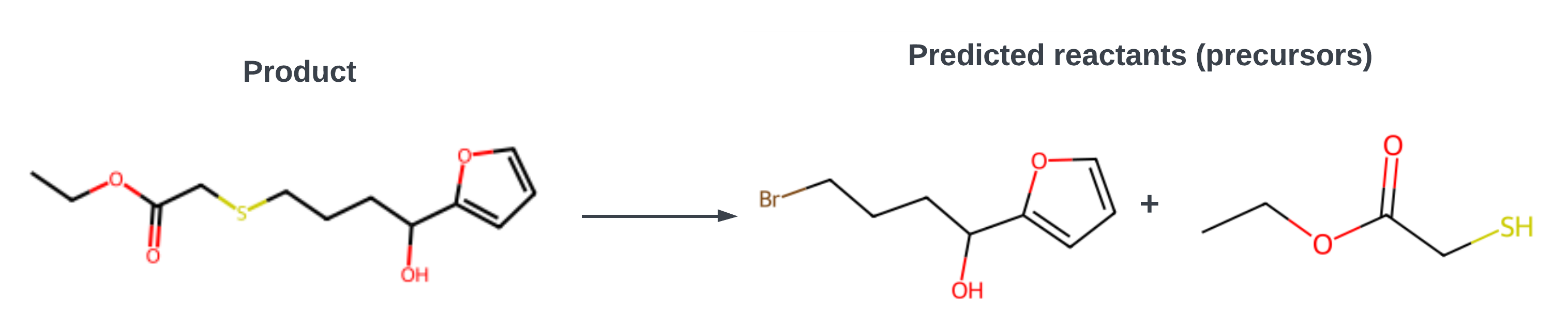}
        \caption{An example top-1 prediction to study the transformer attention map}
        \label{fig:attention-rkn2}
     \end{subfigure}\\
     \begin{subfigure}[b]{0.85\textwidth}
        \centering
        \includegraphics[width=\textwidth]{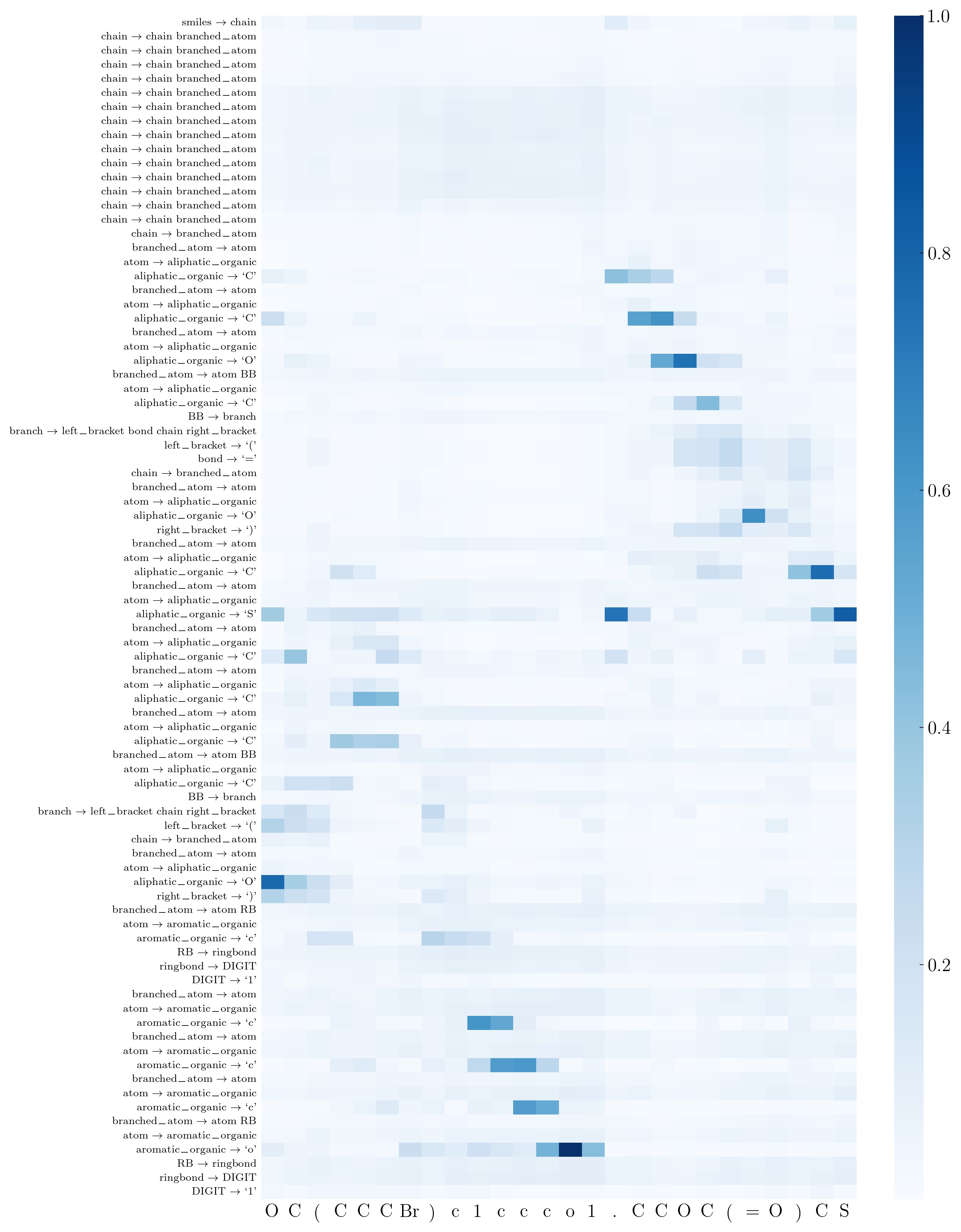}
        \caption{Average attention scores for top-1 prediction on the retrosynthesis reaction \texttt{CCOC(=O)CSCCCC(O)c1ccco1} $\to$ \texttt{OC(CCCBr)c1ccco1.CCOC(=O)CS} extracted from the transformer decoder cross-attention sublayer}
        \label{fig:heatmap2}
     \end{subfigure}
    \caption{Molecular attention map for the SMILES grammar tree of the product and SMILES strings of the reactants (precursors) for an example reaction}
    \label{fig:attn_heatmap2}
\end{figure}

\begin{figure}[H]
    \centering
    \begin{subfigure}[b]{0.8\textwidth}
        \centering
        \includegraphics[width=0.8\textwidth]{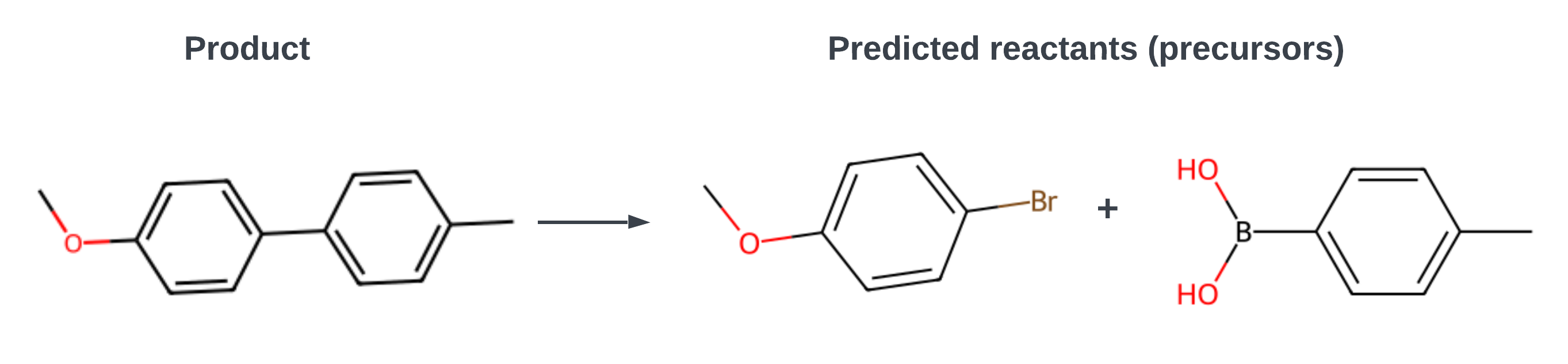}
        \caption{An example top-1 prediction to study the transformer attention map}
        \label{fig:attention-rkn3}
     \end{subfigure}\\
     \begin{subfigure}[b]{0.82\textwidth}
        \centering
        \includegraphics[width=\textwidth]{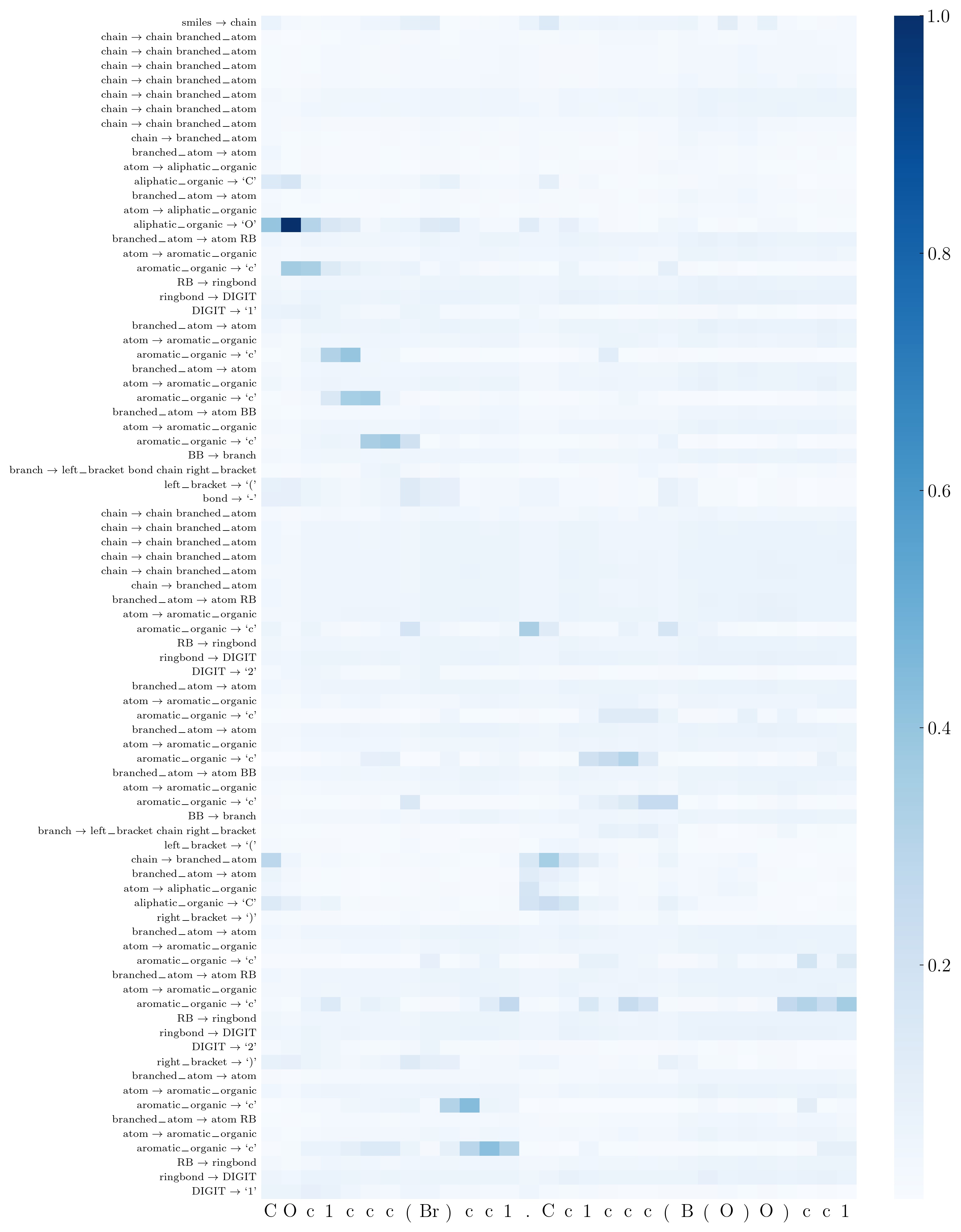}
        \caption{Average attention scores for top-1 prediction on the retrosynthesis reaction \texttt{COc1ccc(-c2ccc(C)cc2)cc1} $\to$ \texttt{COc1ccc(Br)cc1.Cc1ccc(B(O)O)cc1} extracted from the transformer decoder cross-attention sublayer}
        \label{fig:heatmap3}
     \end{subfigure}
    \caption{Molecular attention map for the SMILES grammar tree of the product and SMILES strings of the reactants (precursors) for an example reaction}
    \label{fig:attn_heatmap3}
\end{figure}

We compute the attention map for a more complex reaction \texttt{COc1ccc(-c2ccc(C)cc2)cc1} \\
$\to$ \texttt{COc1ccc(Br)cc1.Cc1ccc(B(O)O)cc1} in Figure \ref{fig:attn_heatmap2} and \texttt{CCOC(=O)CSCCCC(O)c1ccco1} \\
$\to$ \texttt{OC(CCCBr)c1ccco1.CCOC(=O)CS} in Figure \ref{fig:attn_heatmap3}. 

\newpage

\printbibliography







\end{document}